\documentclass[letterpaper,10pt,journal,compsoc]{IEEEtran}

\usepackage{amsmath,amsfonts}
\usepackage{adjustbox}
\usepackage{multirow}
\usepackage{booktabs}
\usepackage[caption=false,font=footnotesize]{subfig}
\usepackage{textcomp}
\usepackage{stfloats}
\usepackage{url}
\usepackage{graphicx}
\usepackage{xcolor}
\usepackage{colortbl}
\usepackage{cite}
\usepackage{enumitem}
\usepackage[most]{tcolorbox}
\usepackage{multicol}

\definecolor{tableblue}{RGB}{239,246,252}

\newcounter{promptbox}
\tcbset{
  mypromptbox/.style={
    enhanced,
    colback=gray!5,
    colframe=red!40!black,
    fonttitle=\bfseries,
    coltitle=white,
    colbacktitle=red!70!black,
    boxrule=1pt,
    arc=4pt,
    outer arc=4pt,
    left=7pt,
    right=7pt,
    top=7pt,
    bottom=7pt,
    toptitle=3pt,
    bottomtitle=3pt
  }
}

\begin{document}

\title{
Learning Interaction between Image and Layout Priors for Joint Image-Layout Generation in Design Templates
}

\author{Shirong Yang, Bo Yang, and Ying Cao%
\thanks{Shirong Yang, Bo Yang, and Ying Cao are with the School of Information Science and Technology, ShanghaiTech University, 393 Middle Huaxia Road, Pudong New Area, Shanghai, China (e-mail: yangshr2024@shanghaitech.edu.cn; borisyang326@gmail.com; caoying59@gmail.com).}%
\thanks{Corresponding author: Ying Cao.}}

\markboth{IEEE Transactions on Visualization and Computer Graphics}{Yang \MakeLowercase{\textit{et al.}}: InterIL}

\maketitle

\begin{abstract}
In this paper, we address the problem of graphic design template creation, which generates a background image and a layout of foreground elements over the background to form a harmonious composition from an input text.
Prior work on graphic design generation mostly adopts a \textit{sequential} paradigm, where design elements are generated sequentially. We argue that such a sequential scheme falls short of faithfully capturing the dependency between the background and layout (and thus the joint image-layout distribution), which limits the quality of generated design templates.
To overcome this limitation, we propose a model, \textit{InterIL}, which \textit{jointly} generates the two modalities\textemdash background image and layout\textemdash{}in a single generative process. The novel design of our joint model connects the backbones of pretrained image and layout diffusion models with a learnable communication module to explicitly model bidirectional image-layout interaction. During training, the image and layout backbones are frozen to maintain and leverage the vast pretrained single-modality prior knowledge, while only the communication module is updated, so that the model can focus on learning image-layout interaction and thereby better capture the joint image-layout distribution for improved composition harmony.
Our model has no design-specific inductive bias, which allows it to better preserve the original characteristics of realistic designs. We further introduce a test-time guidance strategy to enable users to impose their specific preferences on generated results.
Our experiments show that, compared with prior approaches, our model can generate significantly better results in terms of image, layout and image-layout harmonization, producing outputs closer to real samples. We also demonstrate the flexibility of our model in enforcing user preferences
at inference without retraining.
\end{abstract}

\begin{IEEEkeywords}
Design template generation, diffusion models, layout generation.
\end{IEEEkeywords}

\begin{figure*}[!t]
\centering
\includegraphics[width=\textwidth]{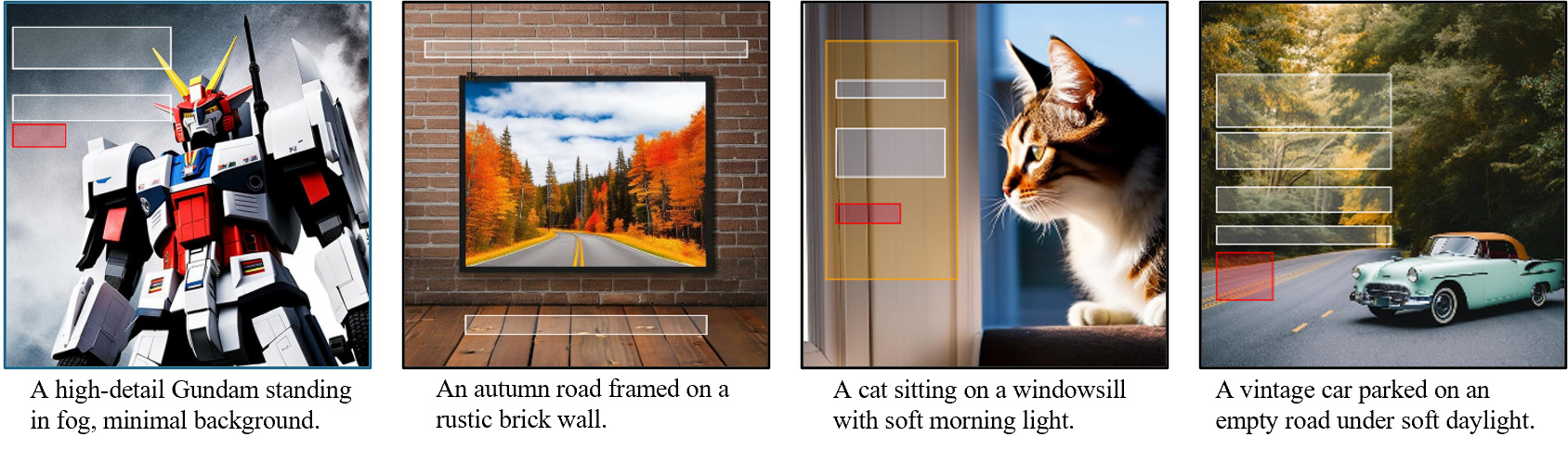}
\caption{Generated design templates by our model given text inputs. Each template specifies the overall theme and structure of the design, consisting of a background image and a layout of elements of different types: text (white), underlay (yellow), and button (red). The input text is shown below each generated template.}

\label{fig:teaser}
\end{figure*}

\section{Introduction}
\label{sec:intro}

\IEEEPARstart{W}{hen} creating graphic designs, designers usually choose and start from pre-existing design templates, instead of designing from scratch, and iteratively refine the templates towards the final full designs. Design templates are of great value to designers in practice, which can provide inspirational ideas and help speed up the design process. However, creating good design templates is non-trivial, which often takes substantial manual effort and necessitates considerable design expertise.

In this paper, we aim to automate the creation of graphic design templates from input textual descriptions. Ideally, design templates should lie midway between high-level requirements and finalized designs, serving as a \textit{mid-level abstraction} of complete designs: they are more concrete than high-level intentions with overall theme and structure to serve as good starting points, while being less complete than full designs with only a partial set of element attributes to leave room for creativity. In this work, we focus on a specific form of design templates, which is composed of a background image and a layout (spatial arrangement) of foreground elements over the background~\cite{weng2024desigen}.

Despite remarkable progress in generative models for graphic design, previous methods mostly follow a \textit{sequential} generation paradigm, where design elements are created one after another. Cascaded pipelines~\cite{jia2023cole,inoue2024opencole}
execute a series of sub-tasks (e.g., background generation, object generation and typography generation) in sequence. The causal nature of the autoregressive design generation models~\cite{lin2025elements,qu2025igd,yang2025order}
imposes a dependency order between generated elements. A recent work on design template generation~\cite{weng2024desigen} takes a sequential two-stage approach, where a background is first generated, followed by generating a layout conditioned on the background. However, in design templates, the dependency between background and layout is \textit{bidirectional}\textemdash background generation should take the layout into account (e.g., to allocate appropriate space for foreground element placement); layout generation should be aware of the background (e.g., to not occlude important background regions). As a result, the existing methods are limited in modeling desired background-layout interaction, which is crucial for harmonious compositions, and thus lead to degraded generation quality (e.g., visual conflict between background and layout).

To mitigate this issue, we propose \textit{InterIL}, a latent diffusion model for design template generation, which enables \textit{joint} generation of images and layouts in a \textit{single} framework by learning
image-layout interaction.
Specifically, we first pretrain a latent diffusion model for unconditional layout generation as a \textit{layout prior}, and leverage a pretrained text-to-image latent diffusion model~\cite{rombach2022high} as an \textit{image prior}. Each of the two priors alone can generate high-quality samples in its own domain (image or layout). We then construct the denoising network of InterIL by combining the pretrained backbones of the two priors and connecting them with a communication module dedicated to explicitly learning bidirectional image-layout interaction. When training InterIL,
we keep the pretrained weights of the two backbones fixed, and only optimize the communication module. This allows us to focus learning on capturing
image-layout interaction patterns, as the prior knowledge of generating realistic images and layouts is already available in the two backbones. With the learned communication module, InterIL can generate a holistic design template with a single denoising process, through which the image and layout backbones frequently communicate with each other to denoise noisy image and layout latents jointly.
Due to the joint generation strategy and the explicit image-layout interaction learning, our model is able to capture the joint image-layout distribution more faithfully than existing sequential methods.
Moreover, unlike the previous work~\cite{weng2024desigen}, our model refrains from using any design-related inductive bias, and therefore better fits the training data distribution, reproducing the unique characteristics involved in realistic designs.
Our model also offers flexibility in deviating from realistic design characteristics and shifting generations towards user-preferred patterns (e.g., minimizing occlusion of salient background areas), through a proposed training-free guidance strategy.

To validate the effectiveness of our InterIL model, we introduce two evaluation metrics, \textit{TemplateFID} and \textit{TemplateCLIP}, which respectively assess the visual quality and text adherence of generated templates. We conduct experiments on the Web-design dataset~\cite{weng2024desigen}, comparing InterIL with alternative approaches. Our model exhibits dramatically improved performance compared to the other methods, generating high-quality, visually harmonious design templates (see Figure~\ref{fig:teaser}).
We also showcase how our proposed guidance technique can be applied at inference time to allow users to control generated templates through their expressed preferences, including reducing foreground-element occlusion of salient background regions and improving text readability.

Our paper makes the following contributions:
\begin{itemize}
\item A new \textit{joint} image-layout generation paradigm for graphic design template generation, which differs from the existing \textit{sequential} generation scheme.
\item A diffusion-based design template generation framework, which synthesizes images and layouts \textit{simultaneously} in a \textit{single} generation process by learning the interaction between the pretrained image and layout priors, which is in contrast to the \textit{sequential}, \textit{multi-stage} framework used in previous works.
\item An
inference-time guidance framework that
can accommodate different user preferences on generated results without any model retraining.
\end{itemize}

\section{Related Work}
\label{sec:related_work}

\subsection{Graphic Layout Generation}
Graphic layout generation, which involves creating the spatial arrangement of elements on a canvas,
has been extensively studied~\cite{jacobs2003adaptive,odonovan2014learning,pang2016directing,qiao2023design}
Recent works have approached layout generation using various deep generative modeling frameworks, including Generative Adversarial Networks (GANs)~\cite{li2019layoutgan,zheng2019content,Kikuchi2021}, Variational Autoencoders (VAEs)~\cite{jyothi2019layoutvae,arroyo2021variational}, Transformers~\cite{gupta2021layouttransformer, horita2024retrieval, jiang2023layoutformer++}, diffusion models~\cite{inoue2023layoutdm, zhang2023layoutdiffusion,cheng2023play}, and flow-based models~\cite{guerreiro2024layoutflow}.
Some works also show the effectiveness of large language models (LLMs) in solving the layout generation task based on the layout-related knowledge that LLMs have acquired during pretraining~\cite{lin2024layoutprompter}.
Constrained layout generation has also been investigated, which imposes various user constraints on element attributes and relationships to control generated layouts~\cite{lee2020neural, Kikuchi2021, jiang2023layoutformer++}. Several recent methods investigate graphic design composition, in which a set of multimodal elements (images and texts) is used as conditioning information and composed into a cohesive design~\cite{graphist2023hlg, shabani2024visual, zhang2025creatidesign, lin2025elements}. Our work is related to a branch of methods on content-aware layout generation~\cite{zheng2019content},
which generates layouts conditioned on a background image~\cite{zhou2022composition,cao2022geometry,horita2024retrieval,seol2024posterllama,wang2025sega}.
However, instead of tackling background-conditioned layout generation solely, we aim to generate both a layout and a background image to constitute a design template.

\subsection{Graphic Design Generation}
Recently, there has been a rising interest in developing models for graphic design generation.
For example, CanvasVAE~\cite{yamaguchi2021canvasvae} trains a VAE to generate graphic designs represented as sets of canvas and element attributes. GOL~\cite{yang2025order} shows that learned element order can improve the performance of generative models that predict a sequence of element attribute tokens.
Several methods have been proposed to automate graphic design generation from textual design intentions. One class of methods directly fine-tunes pretrained text-to-image diffusion models for text-to-design generation, producing highly aesthetic and coherent design images~\cite{wang2025designdiffusion,chen2025postercraft}. Another class of methods tries to generate layered graphic designs, composed of multiple image and text layers, using a cascade of task-specific models~\cite{jia2023cole, inoue2024opencole}, or a multimodal large language model with an image-generation capability~\cite{qu2025igd}.

The design template generation problem that we focus on can be viewed as a specialization of design generation that considers a subset of element attributes, including background image and element layout.
A recent work, Desigen~\cite{weng2024desigen}, presents a solution to the problem. Desigen takes a two-stage approach that generates the background and layout sequentially using two separate models, and captures the dependency between them by simply conditioning one model on the final output of the other. This method suffers from two major shortcomings: first, the communication between the image and layout models is limited, which restricts its ability to capture interaction between image and layout; second, it explicitly introduces an inductive bias towards minimizing occlusion of salient background regions, which causes the distribution of generated samples to drift from the data distribution, washing out distinctive patterns of real samples from the model outputs.
In contrast, our model generates a background image and a layout jointly in a single diffusion model, which focuses on learning image-layout interaction to achieve more coherent image-layout composition. In addition, our model imposes no design-specific inductive bias, thereby better maintaining the distinctive features of training samples in generated design templates.

\section{Method}
\label{sec:method}

Given a text description $\mathcal{P}$,
our goal is to generate a design template $X^D$. Each design template consists of a background image $X^I$ and a layout $X^L$, where the layout comprises a set of elements defined by their categories and bounding box coordinates.
To this end, we aim to learn a joint distribution $p(X^I, X^L|\mathcal{P})$ of background images $X^I$ and layouts $X^L$ conditioned on $\mathcal{P}$, from which we can sample $X^I$ and $X^L$, and compose them into a coherent design template. The key idea of our method is to 1) train two expressive diffusion priors on image and layout domains (to ensure the high-quality generation of background images and layouts), and 2) learn the interaction between the two priors in a single joint diffusion model (to ensure a harmonious composition). This leads to a unified model, \textit{InterIL}, for joint generation of background images and layouts from text inputs.

\begin{figure*}[t]
    \centering
    \includegraphics[width=0.95\linewidth]{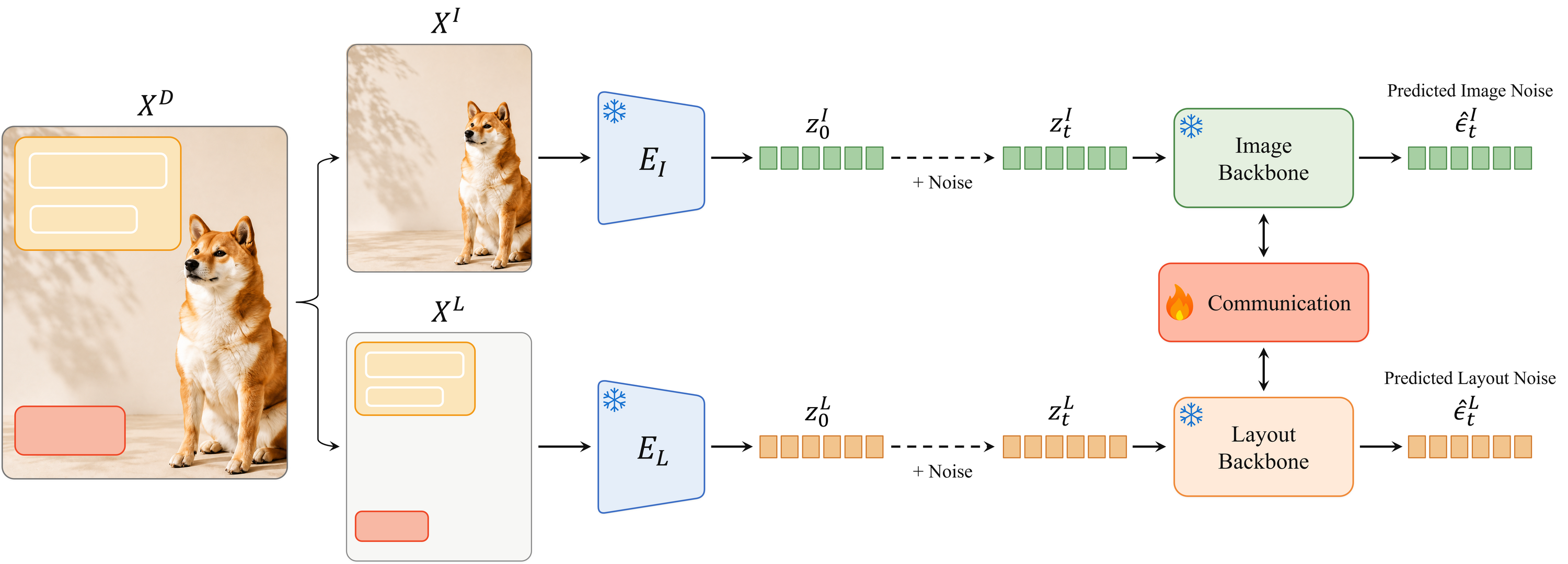}
    \caption{Overview of our joint model. Our model jointly denoises the noisy image and layout latents with an architecture that connects the pretrained, fixed backbones of image and layout diffusion models with a learnable communication module for explicitly learning bidirectional image-layout interaction.
    }

    \label{fig:main-pipeline}
\end{figure*}

\subsection{Domain-specific Priors}
\label{subsec:domain-spec-prior}

We aim to build a conditional image prior $p(X^I|\mathcal{P})$ that can generate images from text prompts, and a layout prior $p(X^L)$ for generating high-quality layouts.

\textbf{Image Prior.} For $p(X^I|\mathcal{P})$, we employ Stable Diffusion (\textit{SD})~\cite{rombach2022high}, a large-scale latent diffusion model (LDM) for text-to-image generation. SD trains a variational autoencoder (VAE) with an image encoder $\mathcal{E}_{\text{I}}$ and an image decoder $\mathcal{D}_{\text{I}}$. The encoder $\mathcal{E}_{\text{I}}$ maps an RGB image $X^I \in \mathbb{R}^{H \times W \times 3}$ to a spatial latent representation $z_0^I = \mathcal{E}_{\text{I}}(X^I) \in \mathbb{R}^{h \times w \times d^I}$ (an $h \times w$ grid of embeddings, each of dimensionality $d^I$). Then, a diffusion model is learned over latents instead of pixels. For image generation, a latent $\tilde{z}_0^I$ is sampled from the diffusion model and then decoded by the decoder $\mathcal{D}_{\text{I}}$ into an image $\tilde{X}^I = \mathcal{D}_{\text{I}}(\tilde{z}_0^I)$.
To adapt the pretrained SD to the task of generating background images in graphic designs, we fine-tune it on a dataset of background images associated with text descriptions to obtain the adapted denoising network $\epsilon_{\theta^*}$, which we refer to as the \textit{image backbone} in our joint model introduced later.

\textbf{Layout Prior.}
For $p(X^L)$, we train another LDM from scratch on a layout dataset. Following prior work on layout generation~\cite{gupta2021layouttransformer, jiang2023layoutformer++, inoue2023layoutdm}, we define a layout element as a bounding box with 5 attributes, including category, left coordinate, top coordinate, width, and height. After discretizing the continuous attributes into bins using the k-means algorithm, an element is formatted as a sequence of 5 attribute tokens, and a layout is the concatenation of the element sequences. We first train a VAE to project layouts into a latent space. The Transformer-based VAE encoder $\mathcal{E}_{\text{L}}$ encodes a layout $X^L$ into a latent representation $z_0^L = \mathcal{E}_{\text{L}}(X^L) \in \mathbb{R}^{N \times d^L}$, where $N$ is the number of elements in the layout and $d^L$ is the embedding dimensionality. We then train a diffusion transformer (DiT)~\cite{peebles2023scalable} over the latent space. A layout can be generated by sampling $\tilde{z}_0^L$ from the DiT, followed by feeding it into the Transformer-based VAE decoder $\mathcal{D}_{\text{L}}$: $\tilde{X}^L = \mathcal{D}_{\text{L}}(\tilde{z}_0^L)$. We refer to the trained Transformer-based denoising network $\epsilon_{\phi^*}$ of the DiT as the \textit{layout backbone} in our subsequent joint model.

\subsection{Joint Model}
Given the two pretrained priors that operate independently, we merge them into a single diffusion model to simultaneously generate background images and layouts. As illustrated in Figure~\ref{fig:main-pipeline}, the core design of the joint model's denoising network is to put the \textit{pretrained} and \textit{frozen} image and layout backbones together to jointly denoise noisy image and layout latents, while connecting the backbones using a \textit{learnable} communication module to capture image-layout interaction. Such a design has two primary benefits: first, by locking the weights of the backbones, we can retain the pretrained knowledge in them to effectively leverage their original capabilities to ensure the quality of generated images and layouts; second, as the prior knowledge of how to generate images and layouts independently is already available in the two pretrained backbones, our joint model, when trained on design template datasets, can focus on learning
interaction between image and layout, thus improving composition harmony. Furthermore, the two backbones can communicate over a large number of denoising steps \textit{during} the generation process of images and layouts. This is different from the sequential iterative refinement (e.g., in Desigen), where the image (layout) is generated \textit{after} the generation of the layout (image).
Consequently, our joint model can better model image-layout interaction than prior work.

To learn our model, we train a joint denoising network:
\begin{align}
[\hat{\epsilon}_t^I, \hat{\epsilon}_t^L] =
\epsilon_{\psi, \theta^*, \phi^*}(z_t^I, z_t^L, t, \mathcal{P}),
\end{align}
where $\hat{\epsilon}_t^I, \hat{\epsilon}_t^L$ are predicted noises. The training is performed by optimizing the following objective:
\begin{align}
    \label{eq:loss}
    \mathcal{L} = \mathbb{E}_{z^I_0, z^L_0, t, \epsilon^I_t, \epsilon^L_t} \Big[ \lambda_I\| \epsilon_t^I - \hat{\epsilon}_t^I \|_2^2 + \lambda_L\| \epsilon_t^L - \hat{\epsilon}_t^L \|_2^2\Big],
\end{align}
where $t \sim [1, T]$ (T is the number of diffusion timesteps), $\epsilon^I_t \sim \mathcal{N}(\mathbf{0}, \mathbf{I})$, and $\epsilon^L_t \sim \mathcal{N}(\mathbf{0}, \mathbf{I})$. Our implementation sets $\lambda_L = 3$ and $\lambda_I = 1$. $\psi$ contains the communication module's trainable weights.

\subsection{Communication Module}

\begin{figure*}[t]
    \centering
    \includegraphics[width=\textwidth]{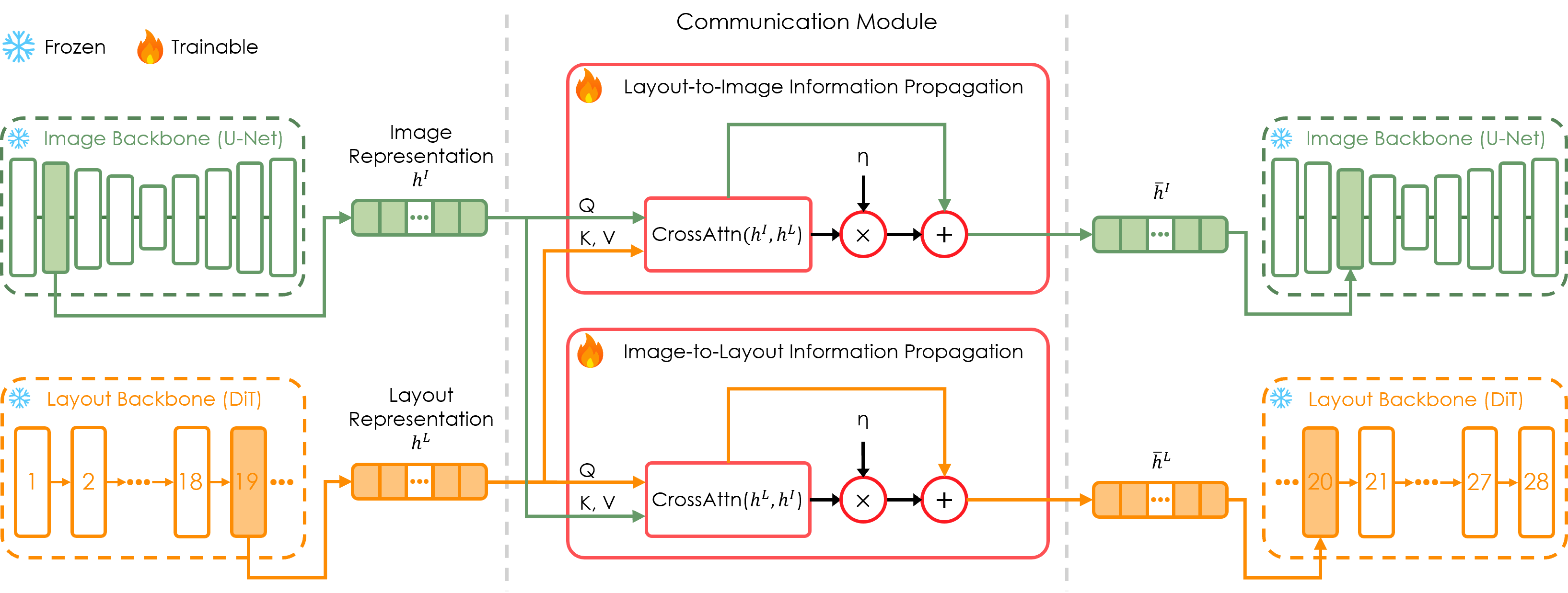}
    \caption{
    The communication module enables bidirectional information exchange between the image backbone and layout backbone by augmenting the intermediate representation of each backbone with that of the other backbone through cross-attention.
    }

    \label{fig:communication_detail}
\end{figure*}

As illustrated in Figure~\ref{fig:communication_detail}, the communication module is designed to facilitate bidirectional information exchange between the image and layout backbones, while preserving their pretrained knowledge as much as possible. Let $h^I \in \mathbb{R}^{M \times d}$ be the flattened intermediate representation of the U-Net-based image backbone (from the second encoder block, the third encoder block or the middle block),
and $h^L \in \mathbb{R}^{N \times d}$ be the output representation of the 19th DiT block of the layout backbone. The communication module augments $h^I$ and $h^L$ as:
\begin{align}
    [\bar{h}^I, \bar{h}^L] = \operatorname{COMM}(h^I, h^L).
\end{align}
The augmented representations $\bar{h}^I, \bar{h}^L$ are then fed into the subsequent blocks in the image and layout backbones, respectively.
More specifically, to obtain $\bar{h}^I$, the operations can be written as:
\begin{align}
    \label{eq:comm1}
    \bar{h}^I = h^I + \eta \cdot \operatorname{CrossAttn}(h^I, h^L).
\end{align}
The cross-attention can model image-layout interaction, propagating layout information from $h^L$ to $h^I$.
$\eta$ is set to 1 during training and can be varied during inference to control when the communication is enabled over the denoising process. Similarly, $\bar{h}^L$ is obtained by:
\begin{align}
    \label{eq:comm2}
    \bar{h}^L = h^L + \eta \cdot \operatorname{CrossAttn}(h^L, h^I).
\end{align}

In this work, we opt to make the design of the communication module simple and effective (verified in Section \ref{exp:comm}), and leave the exploration of more sophisticated architectural designs for future work.

\begin{figure*}[t]
    \centering
    \includegraphics[width=0.9\textwidth]{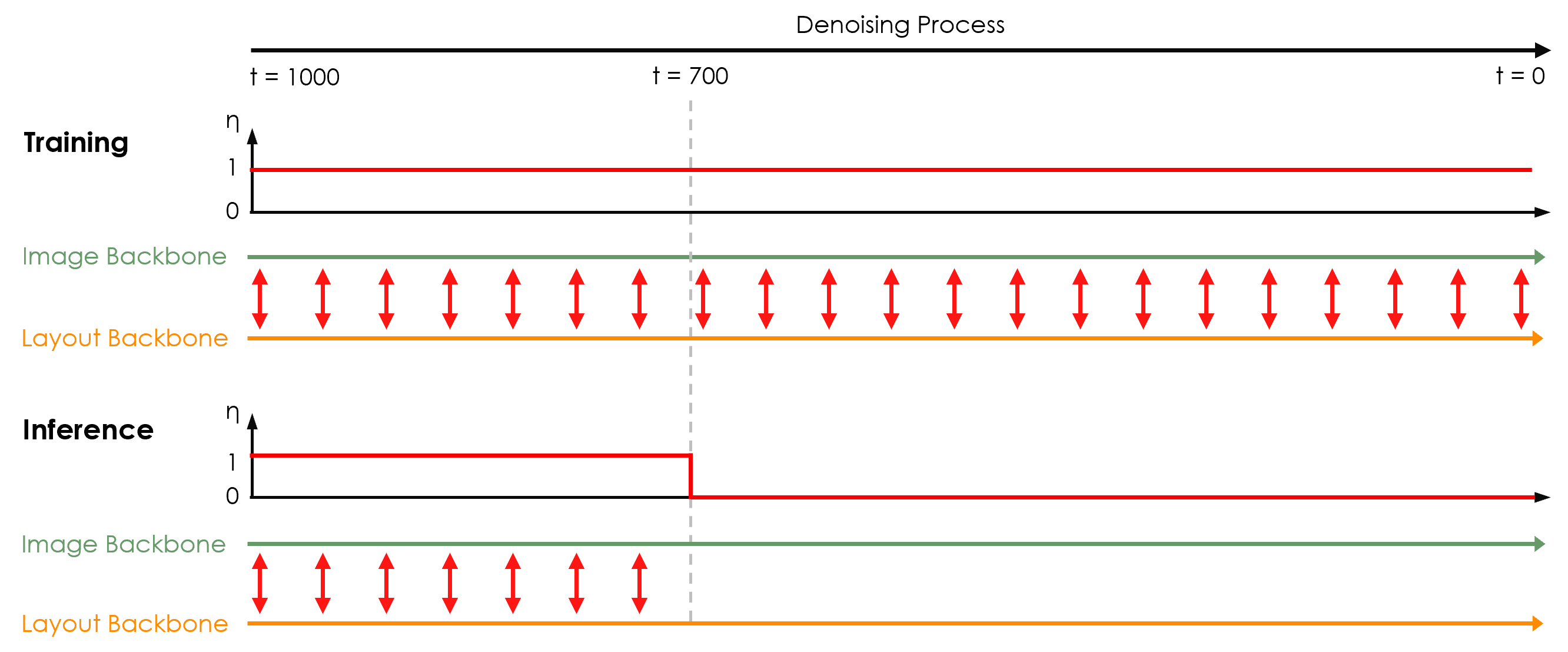}
    \caption{
    Scheduling the communication between the image and layout backbones during training and inference by setting the value of $\eta$.
    During training,  the communication is enabled by setting $\eta = 1$ at all diffusion time steps. During inference, the communication is only enabled for the first $30\%$ of the denoising process where $\eta = 1$, and is turned off for the remaining denoising steps where $\eta = 0$.
    }

    \label{fig:eta_schedule}
\end{figure*}

When training the model, we set $\eta=1$ in Equation~\ref{eq:comm1} and~\ref{eq:comm2}, enabling the communication module for all diffusion timesteps so that this module can be sufficiently updated. However, our early experiments show that using $\eta=1$ throughout the entire denoising process at test time leads to unsatisfactory layout generation quality. To mitigate this issue, inspired by GLIGEN~\cite{li2023gligen}, we dynamically adjust the value of $\eta$ during the sampling process to improve visual quality.
The denoising process of diffusion models generates coarse-grained features (e.g., global structures and colors) at early steps, and then generates perceptually important content and fine-grained details at later steps. Therefore, as illustrated in Figure~\ref{fig:eta_schedule}, we opt to turn on the communication module ($\eta = 1$) during the first $30\%$ of the denoising process, where the image backbone can use \textit{external} knowledge from the layout backbone to generate a rough spatial arrangement of objects in the image that harmonizes with the layout, and vice versa for the layout backbone. In the remaining denoising steps, we disable the communication module ($\eta = 0$), allowing the two backbones to rely upon their \textit{internal} knowledge to generate high-quality images and layouts.

\subsection{Preference-based Guidance}
Our model uses no design-specific inductive bias, thereby enjoying the benefit of preserving the distinctive design patterns in realistic (training) design samples. However, in some cases, users may prefer the occurrence of specific design characteristics in generated templates (e.g., minimal occlusion of important background regions). To address this, we propose a guidance technique that is applied at inference to impose preference-based constraints on generated results.
Specifically, we guide the layout generation process by adjusting the noise prediction $\hat{\epsilon}_t^L$ of the layout backbone:
\begin{align}
    \hat{\epsilon}_t^L \leftarrow \hat{\epsilon}_t^L - s \sqrt{1 - \bar{\alpha}_t}\,\nabla_{z_t^L}\mathcal{L}_{\text{guide}}\big(\hat{\epsilon}_t^I, \hat{\epsilon}_t^L\big),
\end{align}
where $s$ is the guidance scale, $\bar{\alpha}_t$ is the multiplication of the forward-process noise variances from $1$ to $t$, and $\mathcal{L}_{\text{guide}} = \sum_{k=1}^K \lambda_k \mathcal{L}_k$ is a weighted combination of differentiable objectives that represent different user preferences.
$[\hat{\epsilon}_t^I, \hat{\epsilon}_t^L] = \epsilon_{\psi^*, \theta^*, \phi^*}(z_t^I, z_t^L, t, \mathcal{P})$, where $\psi^*$ denotes the trained weights obtained by minimizing Equation~\ref{eq:loss}.

We instantiate this guidance technique with two concrete objectives as follows.
The first objective represents preference for minimizing occlusion, discouraging foreground elements from occluding salient background regions. Given the predicted image noise $\hat{\epsilon}_t^I$, we compute a clean image latent $\hat{z}_0^I$ analytically using the forward process marginal distribution $q(z_t^I|z_0^I)$, decode it into an image $\mathcal{D}_{\text{I}}(\hat{z}_0^I)$, and compute a saliency map $S = f_\text{sal}(\mathcal{D}_{\text{I}}(\hat{z}_0^I))$ using an off-the-shelf saliency detector $f_\text{sal}$. We then compute a clean layout latent $\hat{z}_0^L$ from the predicted layout noise $\hat{\epsilon}_t^L$ with $q(z_t^L|z_0^L)$, and decode it into a layout $\mathcal{D}_{\text{L}}(\hat{z}_0^L)$. Let $\{\mathbf{b}_i\}_{i=1}^B$ be the decoded element bounding boxes. The objective is defined as:
\begin{align}
\mathcal{L}_\text{occ}
= \frac{1}{B}\sum_{i=1}^B \frac{1}{A_i} \sum_{p} M_{\mathbf{b}_i}(p)\, S(p),
\end{align}
where $M_{\mathbf{b}_i}$ is a soft mask for $\mathbf{b}_i$, $A_i$ is the area of $\mathbf{b}_i$, and $p$ indexes spatial positions.
To compute $M_{\mathbf{b}_i}$ in a way that ensures that $\mathcal{L}_\text{occ}$ is differentiable, we convert discrete bounding box parameters to continuous ones by computing each parameter as a weighted sum of quantization bin centers, weighted by predicted probabilities. Given a bounding box with continuous parameters $\mathbf{b}_i = (x_i, y_i, w_i, h_i)$, we derive the corner coordinates $(x^l_i, y^t_i)$ and $(x^r_i, y^b_i)$, and then compute $M_{\mathbf{b}_i}$ as:
\begin{equation}
\begin{split}
    M_{\mathbf{b}_i}(x, y) =  \sigma & (\lambda (x - x^l_i)) \times \sigma(\lambda (x^r_i - x)) \\
    & \times \sigma(\lambda (y - y^t_i)) \times \sigma(\lambda (y^b_i - y)),
\end{split}
\end{equation}
where $\sigma(\cdot)$ is the sigmoid function and $\lambda = 40$.
We refer to guidance using this objective as \textit{occlusion-aware guidance} (OAG).

The second objective represents preference for improved text readability, which encourages text elements to lie on flat
background regions. From the decoded image $\tilde{I} = \mathcal{D}_{\text{I}}(\hat{z}_0^I)$, we construct
a map $C$ that has higher values for positions in $\tilde{I}$ where high-frequency patterns occur:

\begin{align}
    \label{eq:clutter-map}
    C(x,y) = \sqrt{
    \left\| (K_x * \tilde{I})(x,y) \right\|_2^2
    + \left\| (K_y * \tilde{I})(x,y) \right\|_2^2
    + \varepsilon},
\end{align}
where $K_x$ and $K_y$ are the horizontal and vertical Sobel kernels, respectively, $*$ denotes convolution, the $\ell_2$ norm is computed over the RGB channels, and $\varepsilon$ is a small constant for numerical stability.

From the decoded layout $\mathcal{D}_{\text{L}}(\hat{z}_0^L)$, we select
bounding boxes corresponding to text elements, denoted by $\{\mathbf{b}^{\text{text}}_i\}_{i=1}^{B_T}$. The objective is defined as:
\begin{align}
\mathcal{L}_{\text{read}}
= \frac{1}{B_T}\sum_{i=1}^{B_T} \frac{1}{A_i} \sum_{p} M_{\mathbf{b}^{\text{text}}_i}(p)\, C(p),
\end{align}
where $M_{\mathbf{b}^{\text{text}}_i}$ is a differentiable soft mask for $\mathbf{b}^{\text{text}}_i$, which is computed in the same way as $M_{\mathbf{b}_i}$ above. We refer to guidance using this objective as \textit{readability-aware guidance} (RAG).

\section{Experiments}
\label{sec:experiments}

\subsection{Dataset and Implementation Details}
We conduct our experiments on the Web-design dataset~\cite{weng2024desigen}, which consists of 50K web banner designs collected from
online shopping platforms.  We use 41,270 samples (85\%) for training, 2,427 (5\%) for validation, and 4,856 (10\%) for testing.
For pretraining the layout prior, we train its VAE (embedding dimensionality 32) using $\beta$-VAE with $\beta = 5 \times 10^{-4}$. The layout backbone is trained for 1000 epochs with a batch size of 4096. For the image prior, we fine-tune the image backbone for 100 epochs with a learning rate of $1 \times 10^{-5}$ on the background images from the Web-design. Both image and layout models use 1000 diffusion timesteps.
During inference, we use the DDIM sampler for the layout backbone and the DDPM sampler for the image backbone, each using 50 sampling steps. The communication module is enabled at time steps greater than 700 to facilitate information exchange between the layout and image backbones.
Both occlusion-aware guidance and readability-aware guidance are applied between time steps 200 and 700.

\subsection{Compared Methods}
We compare our method with a prior design template generation model, \textit{Desigen}~\cite{weng2024desigen}, which is already trained on the Web-design dataset. Desigen proposes an iterative strategy to refine the generated image and layout. We run this iterative refinement for 3 iterations in the comparison. We also consider a recent text-to-design generation model,  \textit{OpenCOLE}~\cite{inoue2024opencole} as an additional baseline.

\subsection{Evaluation Metrics}

\textbf{Domain-specific Metrics.} Following the evaluation protocol of~\cite{weng2024desigen}, we evaluate the quality of background image and layout separately using the following metrics.
For background image evaluation, we use: \textit{Saliency Ratio}~\cite{weng2024desigen} that measures the proportion of salient regions in an image; \textit{FID}~\cite{heusel2017gans} that measures visual quality and is computed against the testing split; \textit{CLIP Score}~\cite{radford2021learning} that measures text-image alignment. For layout evaluation, we use: \textit{Alignment}~\cite{li2019layoutgan} that measures alignment between layout elements; \textit{Overlap}~\cite{li2019layoutgan} that measures the amount of overlap between layout elements; \textit{Occlusion}~\cite{cao2022geometry} that measures the occlusion of background salient regions by layout elements. As additional layout metrics for comprehensive evaluation, we also include: \textit{LayoutFID}~\cite{Kikuchi2021} that measures overall layout quality and is computed against the testing split; \textit{Readability}~\cite{horita2024retrieval} that measures text readability.

\noindent
\textbf{Holistic Metrics.} We further propose two metrics, \textit{TemplateFID} and \textit{TemplateCLIP}, to evaluate holistic design templates by considering background image and layout jointly. TemplateFID measures how realistic generated templates are, while TemplateCLIP evaluates how well generated templates adhere to the text inputs.

\begin{figure}[t]
\centering
\includegraphics[width=0.48\textwidth]{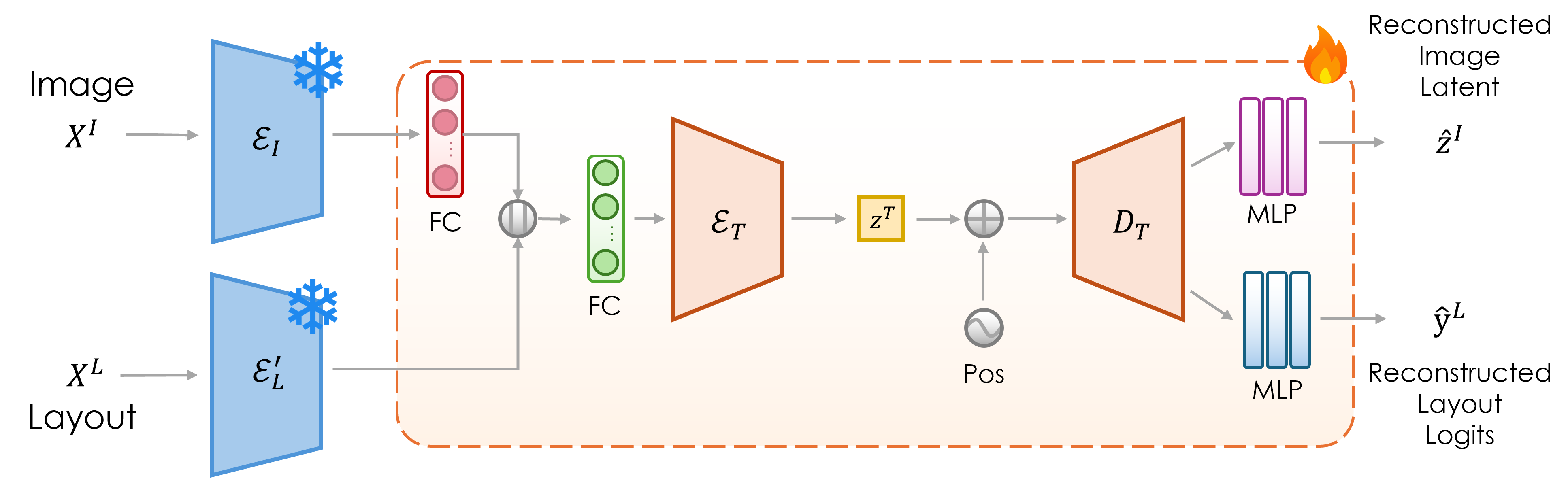}
\caption{
Architecture of the template autoencoder (TemplateAE).
}
\label{fig:dt_ae}
\end{figure}

To compute TemplateFID, we train a template autoencoder (TemplateAE) on a large-scale mixture dataset comprising GenPoster100K~\cite{wang2025sega}, CGL~\cite{ijcai2022p692}, Crello~\cite{yamaguchi2021canvasvae}, and PKU~\cite{Hsu-2023-posterlayout}, deliberately excluding Web-design to avoid evaluation bias. Figure~\ref{fig:dt_ae} illustrates its architecture. Given a background image and a layout, the encoder maps them to a single template embedding. Specifically, an image encoder (the pretrained VAE encoder of Stable Diffusion v1.4) and a layout encoder first map the background image $X^I$ and layout $X^L$ to their respective latent representations $z^I \in \mathbb{R}^{28 \times 28 \times 4}$
and $z^L \in \mathbb{R}^{N \times 32}$, where $N = 7$ is the maximum number of elements in a layout and 32 is the embedding dimensionality. Note that the layout encoder here is not the VAE encoder of the layout prior in InterIL; we train a separate VAE on layouts in the mixture dataset and use its encoder $\mathcal{E}_{\text{L}}^{'}$. Then, $z^I$ is flattened into an embedding of dimensionality 3136 and projected by a fully connected layer to match the dimensionality of $z^L$. The projected $z^I$ is replicated $N$ times to form $\tilde{z}^I \in \mathbb{R}^{N \times 32}$. $\tilde{z}^I$ and $z^L$ are concatenated along the sequence dimension and fed into another fully connected layer, producing a sequence of $2N$ embeddings of dimensionality $512$. The embedding sequence and a learnable embedding (prepended to the sequence) are processed by a template encoder $\mathcal{E}_{\text{T}}$ (a stack of Transformer encoder blocks), and the output of the final block for the learnable embedding is used as the template embedding $z^T \in \mathbb{R}^{512}$. A template decoder $\mathcal{D}_{\text{T}}$ reconstructs the image latent representation and the layout from $z^T$. In particular, the input to $\mathcal{D}_{\text{T}}$ is a sequence of $N$ repeated $z^T$ with sinusoidal positional embeddings added. $\mathcal{D}_{\text{T}}$ (a stack of Transformer encoder blocks) produces a sequence of output embeddings of dimensionality $512$, which are passed through two prediction heads to reconstruct an image latent $\hat{z}^I$ and a layout $\hat{y}^L$ (represented by class logits for each position in the layout sequence). The image head linearly projects each input embedding into dimensionality 3136, and then averages across the sequence dimension to produce a reconstructed image latent $\hat{z}^I$. Given the input embedding for each non-padding layout element, the layout head uses one linear layer to predict class logits for the category and four independent MLPs to predict class logits for each of the four bounding box coordinates. Thus, $\hat{y}^L$ comprises a sequence of non-padding layout elements, where each element occupies five positions and each position has a set of class logits. TemplateAE is trained by optimizing
$\mathcal{L}=\lambda_{\mathrm{img}}\mathcal{L}_{\mathrm{img}}+\lambda_{\mathrm{cls}}\mathcal{L}_{\mathrm{cls}}+\lambda_{\mathrm{bbox}}\mathcal{L}_{\mathrm{bbox}}$. $\mathcal{L}_{\mathrm{img}} = 1-\operatorname{cos}(\hat{z}^I,z^I)$, where $\operatorname{cos}(\cdot,\cdot)$ is cosine similarity. $\mathcal{L}_{\mathrm{cls}}$ computes the mean cross-entropy over the category positions in $\hat{y}^L$, and $\mathcal{L}_{\mathrm{bbox}}$ computes the mean cross-entropy over the bounding box coordinate positions in $\hat{y}^L$. During training, the image encoder and layout encoder are kept frozen, while the template encoder, template decoder, and MLP heads are updated.

We evaluate the quality of the learned template embeddings with a user study involving 104 participants, including 58 design experts and 46 non-experts. We first construct 30 triplets of the form $(x, \tilde{x}_\text{rand}, \tilde{x}_\text{emb})$, where $x$ is a reference template, $\tilde{x}_\text{rand}$ is a randomly selected template from the Web-design dataset, and $\tilde{x}_\text{emb}$ is the most similar template to the reference retrieved from the Web-design dataset based on the learned template embeddings. The subjects were given a triplet and asked to select which of $\tilde{x}_\text{rand}$ and $\tilde{x}_\text{emb}$ is more similar to the reference. The subjects selected the retrieved templates 83.4\% of the time, suggesting that TemplateAE trained on the composed dataset captures important features of design templates and generalizes well to Web-design samples.

To compute TemplateFID, we calculate Fréchet Inception Distance (FID)~\cite{heusel2017gans} between generated and real samples based on the template embeddings. To compute TemplateCLIP, we initialize from the composed-dataset-pretrained TemplateAE encoder and further fine-tune it together with the pretrained SD text encoder using the contrastive learning objective of CLIP~\cite{radford2021learning} on the Web-design train split.

\subsection{Quantitative Results}

\begin{table*}[htbp]
\caption{
Quantitative comparison with Desigen on the Web-design dataset.
“Desigen (n)” indicates applying iterative refinement $n$ times.
For each metric except ones related to CLIP and FID, values closer to that of real data (bottom row) indicate better performance. The best and second-best numerical results are highlighted in \textbf{bold} and \underline{underlined}, respectively.
}
\centering
\begin{adjustbox}{width=\textwidth}
\begin{tabular}{lcccccccccc}
\toprule
\multirow{2}{*}{Method} & \multicolumn{3}{c}{Image} & \multicolumn{5}{c}{Layout} & \multicolumn{2}{c}{Template} \\
\cmidrule(lr){2-4} \cmidrule(lr){5-9} \cmidrule(lr){10-11}
& FID$\downarrow$ & CLIP$\uparrow$ & Saliency Ratio
& LayoutFID$\downarrow$ & Align & Overlap & Occlusion & Readability
& TemplateFID$\downarrow$ & TemplateCLIP$\uparrow$  \\
\midrule
Desigen (0)
& 31.52 & \underline{29.20} & 20.65\%
& 0.56 & \underline{0.35} & \underline{14.41} & \underline{13.47\%} & 11.48\%
& 118.96 & 3.19 \\

Desigen (1)
& 30.84 & 28.87 & 19.29\%
& 0.57 & 0.38 & 15.63 & 13.24\% & 11.56\%
& \underline{108.56} & 3.18 \\

Desigen (2)
& 31.23 & 28.46 & 18.36\%
& 0.51 & 0.37 & 15.52 & 12.70\% & \underline{11.23\%}
& 114.10 & 3.19 \\

Desigen (3)
& \underline{30.79} & 29.04 & \underline{17.67\%}
& \underline{0.48} & 0.37 & 15.23 & 11.45\% & \textbf{10.97\%}
& 116.83 & \underline{3.21} \\

\midrule
\rowcolor{tableblue}
Ours
& \textbf{19.57} & \textbf{29.79} & \textbf{14.56\%}
& \textbf{0.15} & \textbf{0.31} & \textbf{11.98} & \textbf{19.30\%} & 11.41\%
& \textbf{86.63} & \textbf{3.25} \\

\midrule
\textcolor{gray}{Real Data}
& \textcolor{gray}{--} & \textcolor{gray}{27.50} & \textcolor{gray}{14.17\%}
& \textcolor{gray}{--} & \textcolor{gray}{0.31} & \textcolor{gray}{9.34} & \textcolor{gray}{21.14\%} & \textcolor{gray}{8.53\%}
& \textcolor{gray}{--} & \textcolor{gray}{3.39} \\
\bottomrule
\end{tabular}
\end{adjustbox}
\label{tab:comparison_desigen}
\end{table*}

\noindent \textbf{Comparison to Desigen.}
Table~\ref{tab:comparison_desigen} presents the quantitative comparison with Desigen.
We report metrics computed on the test set of the Web-design dataset as a reference, shown as \textit{Real Data} at the bottom of Table~\ref{tab:comparison_desigen}.
For image generation, our method outperforms Desigen across all three metrics, with substantial improvements in FID and saliency ratio. This indicates that our model can generate high-fidelity background images with sufficient empty space for foreground element placement. It should be noted that the CLIP score of real data is lower than those of the other methods, as real background images leave more empty space for foreground elements (lower saliency ratio), which weakens text–image alignment.
For layout generation, our model achieves the best results on 4 out of 5 metrics, and is comparable to Desigen for the readability metric, demonstrating its superior layout generation capability.
Thanks to its explicit learning of image-layout interaction through the communication module, our model is able to generate more harmonious design templates than Desigen, as evidenced by its noticeable improvement in TemplateFID. Furthermore, our model achieves the best TemplateCLIP, showing its strong text adherence.

 As Desigen explicitly enforces occlusion avoidance constraints in its model design, its saliency ratio and occlusion scores decline over refinement iterations, meaning that the salient portion of the generated background image is gradually reduced, and layout elements and salient background regions are moved further apart from each other. While such iterative refinement boosts performance in some metrics (e.g., LayoutFID and readability), a side effect is exposed: the gaps between Desigen's generated samples and real samples, in terms of several aspects (measured by metrics, e.g., alignment, overlap and occlusion), become increasingly larger. The growing gaps imply that the refined results lose some unique design characteristics in training examples, which is undesirable. For example, designers may occasionally make foreground elements partially occlude salient background objects, e.g., for creative or artistic purposes. Excessive occlusion reduction can lead to a lower occlusion score, but may generate unnatural outputs that don't resemble what human designers create. Due to the existence of the aforementioned gaps, Desigen's TemplateFID improves only slightly with the iterative refinement, still significantly lagging behind that of our model.

\begin{table*}[htbp]
\caption{
Quantitative comparison with OpenCOLE on the Web-design dataset.
The text prompts input to our model are augmented by GPT.
For each metric except ones related to CLIP and FID, values closer to that of real data (bottom row) indicate better performance. The best and second-best results are highlighted in \textbf{bold} and \underline{underlined}, respectively.
}
\centering
\begin{adjustbox}{width=\textwidth}
\begin{tabular}{lcccccccccc}
\toprule
\multirow{2}{*}{Method} & \multicolumn{3}{c}{Image} & \multicolumn{5}{c}{Layout} & \multicolumn{2}{c}{Template} \\
\cmidrule(lr){2-4} \cmidrule(lr){5-9} \cmidrule(lr){10-11}
& FID$\downarrow$ & CLIP$\uparrow$ & Saliency Ratio
& LayoutFID$\downarrow$ & Align & Overlap & Occlusion & Readability
& TemplateFID$\downarrow$ & TemplateCLIP$\uparrow$  \\
\midrule

OpenCOLE
& 23.79 & 30.49 & 23.24\%
& 1.63 & 4.20 & \textbf{11.44} & 30.48\% & 11.74\%
& 211.42 & 3.12 \\

\rowcolor{tableblue}
Ours (Prompt Aug.)
& \textbf{18.97} & \textbf{30.94} & \textbf{15.32\%}
& \textbf{0.16} & \textbf{0.32} & 12.06 & \textbf{19.74\%} & \textbf{11.63\%}
& \textbf{85.35} & \textbf{3.28} \\

\midrule
\textcolor{gray}{Real Data}
& \textcolor{gray}{--} & \textcolor{gray}{27.50} & \textcolor{gray}{14.17\%}
& \textcolor{gray}{--} & \textcolor{gray}{0.31} & \textcolor{gray}{9.34} & \textcolor{gray}{21.14\%} & \textcolor{gray}{8.53\%}
& \textcolor{gray}{--} & \textcolor{gray}{3.39} \\
\bottomrule
\end{tabular}
\end{adjustbox}
\label{tab:comparison_opencole}
\end{table*}

\noindent \textbf{Comparison to OpenCOLE.}
OpenCOLE consists of three modules: 1) a design plan generation module (a pretrained large language model) to convert a brief design intention into a detailed design plan; 2) an image generation module (a fine-tuned SD model) that generates an image (i.e., a background image in our problem setting) conditioned on the concatenation of object and background captions from the design plan; 3) a typography generation module (a fine-tuned large multimodal model) that generates text attributes (including the bounding box parameters of texts) based on the design plan and the generated image. For design template generation, OpenCOLE can be seen as a sequential two-stage approach (image generation followed by text layout generation) similar to Desigen, with an additional component (the design plan generation module) to augment the input text prompt. For a fair comparison, we modify the image generation module to use the same SD model as our model, and change the typography generation module to output the layout of elements. The above two modules are fine-tuned on the Web-design dataset. Furthermore, we leverage GPT to turn short text inputs into long, detailed prompts before feeding them into our model. Table~\ref{tab:comparison_opencole} reports the quantitative results of OpenCOLE and our model. It can be seen that our model outperforms OpenCOLE on 9 out of 10 metrics.

\begin{table}[t]
\caption{
GPT-5 evaluation results across four aspects
:
(i) image quality,
(ii) layout quality,
(iii) image--layout harmony,
(iv) text--design relevance.
The best and second-best results are shown in \textbf{bold} and \underline{underlined}, respectively.
}
\centering
\setlength{\tabcolsep}{3pt}
\begin{tabular}{lcccc}
\toprule
Method & (i) & (ii) & (iii) & (iv) \\
\midrule
Desigen
& 6.42 & \underline{6.54} & \underline{7.47} & 6.21 \\

OpenCOLE
& \underline{6.81} & 6.02 & 5.84 & \underline{6.38} \\

\rowcolor{tableblue}
Ours
& \textbf{6.94} & \textbf{7.76} & \textbf{8.07} & \textbf{7.34} \\
\midrule
\textcolor{gray}{Real Data}
& \textcolor{gray}{7.44} & \textcolor{gray}{7.87} & \textcolor{gray}{8.39} & \textcolor{gray}{7.49} \\
\bottomrule
\end{tabular}
\label{tab:gpt5_eval}
\end{table}

\noindent \textbf{GPT-5 Evaluation.} Inspired by recent work on graphic design generation~\cite{jia2023cole,inoue2024opencole}, we also leverage GPT-5 to evaluate design templates generated by Desigen, OpenCOLE, and our model on the Web-design test split. Specifically, GPT-5 assesses the generated results across four aspects: image quality, layout quality, image-layout harmony, and text-design relevance. As summarized in Table~\ref{tab:gpt5_eval}, our model achieves the best results across all four aspects. See the supplementary material for prompt details.

\begin{table}[t]
\caption{
Human preference results. Each number is the percentage of the time that a method is chosen as the best in terms of one aspect.
}
\centering
\setlength{\tabcolsep}{10pt}
\renewcommand{\arraystretch}{1.08}
\begin{tabular}{lccc}
\toprule
{}Aspect & Desigen & OpenCOLE & Ours \\ \midrule
{}Image Quality & 13\% & 19\% & 68\% \\
Layout Quality & 14\% & 8\% & 78\% \\
Image--Layout Harmony & 13\% & 4\% & 83\% \\
Text--Design Relevance & 13\% & 16\% & 71\% \\
\bottomrule
\end{tabular}
\label{tab:human_preference}
\end{table}

\noindent \textbf{Human Evaluation.} We further conduct a human evaluation with 32 participants. For the evaluation, 40 input text prompts from the Web-design test split
are randomly selected. The participants are shown an input text, along with three design templates that are generated from the text by Desigen, OpenCOLE, and our model, respectively, and are presented in randomized order. They are asked to select the best template for each of the four aspects: 1) image quality, 2) layout quality, 3) image--layout harmony, and 4) text--design relevance. The results are shown in Table~\ref{tab:human_preference}. Our model is strongly preferred over Desigen and OpenCOLE on all aspects, especially \textit{image-layout harmony}.

\begin{table}[t]
\caption{
Average inference time per design template on the Web-design test set.
}
\centering
\footnotesize
\setlength{\tabcolsep}{4pt}
\renewcommand{\arraystretch}{1.15}
\begin{tabular}{@{}lc@{}}
\toprule
\textbf{Method} & \textbf{Time} \\
\midrule
Desigen (0) & 7.47 \\
Desigen (1) & 14.85 \\
Desigen (2) & 22.42 \\
Desigen (3) & 29.89 \\
GPT-4o & 10.86 \\
OpenCOLE & 25.80 \\
\midrule
\rowcolor{tableblue}
\textbf{Ours} & \textbf{5.01} \\
Ours + OAG & 8.69 \\
Ours + RAG & 8.47\\
Ours (Prompt Aug.) & 10.14 \\
\bottomrule
\end{tabular}
\label{tab:inference_time}
\end{table}

\noindent \textbf{Computational Efficiency.}
Table~\ref{tab:inference_time} reports the inference time of different methods measured on a single NVIDIA A40 GPU. Due to our single-stage joint image-layout generation scheme, the base model is the most compute-efficient among all the methods. While the iterative refinement of Desigen can improve generation quality with more iterations, it incurs a substantial increase in computational cost\textemdash using 3 refinement iterations increases inference time from 7.47\,s to 29.89\,s. OpenCOLE is the most computationally expensive because its cascaded pipeline relies on multiple large models.
Applying OAG or RAG to our model
introduces only modest additional cost. The guided and prompt-augmented variants of our model remain considerably faster than OpenCOLE and Desigen with iterative refinement.

\subsection{Qualitative Results}

\begin{figure*}[t]
\centering
\includegraphics[width=\textwidth]{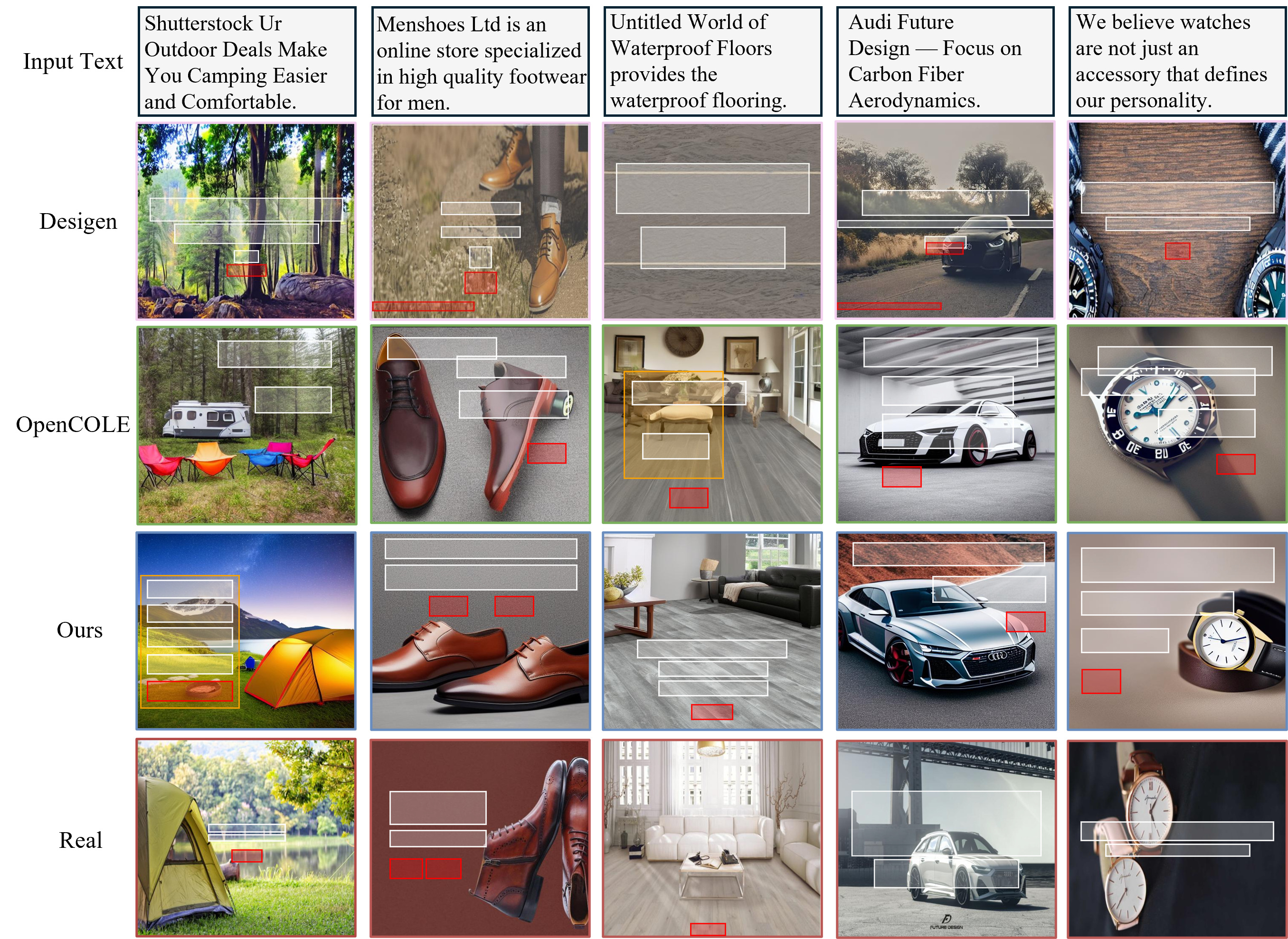}
\caption{
Visual comparison of different methods.
White, orange and red boxes denote text, underlay and button, respectively.
Input texts are shown at the top of each column, and the last row displays the ground truth design templates.
}

\label{fig:qualitative_res}
\end{figure*}

\begin{figure*}[!h]
\centering
\includegraphics[width=\textwidth]{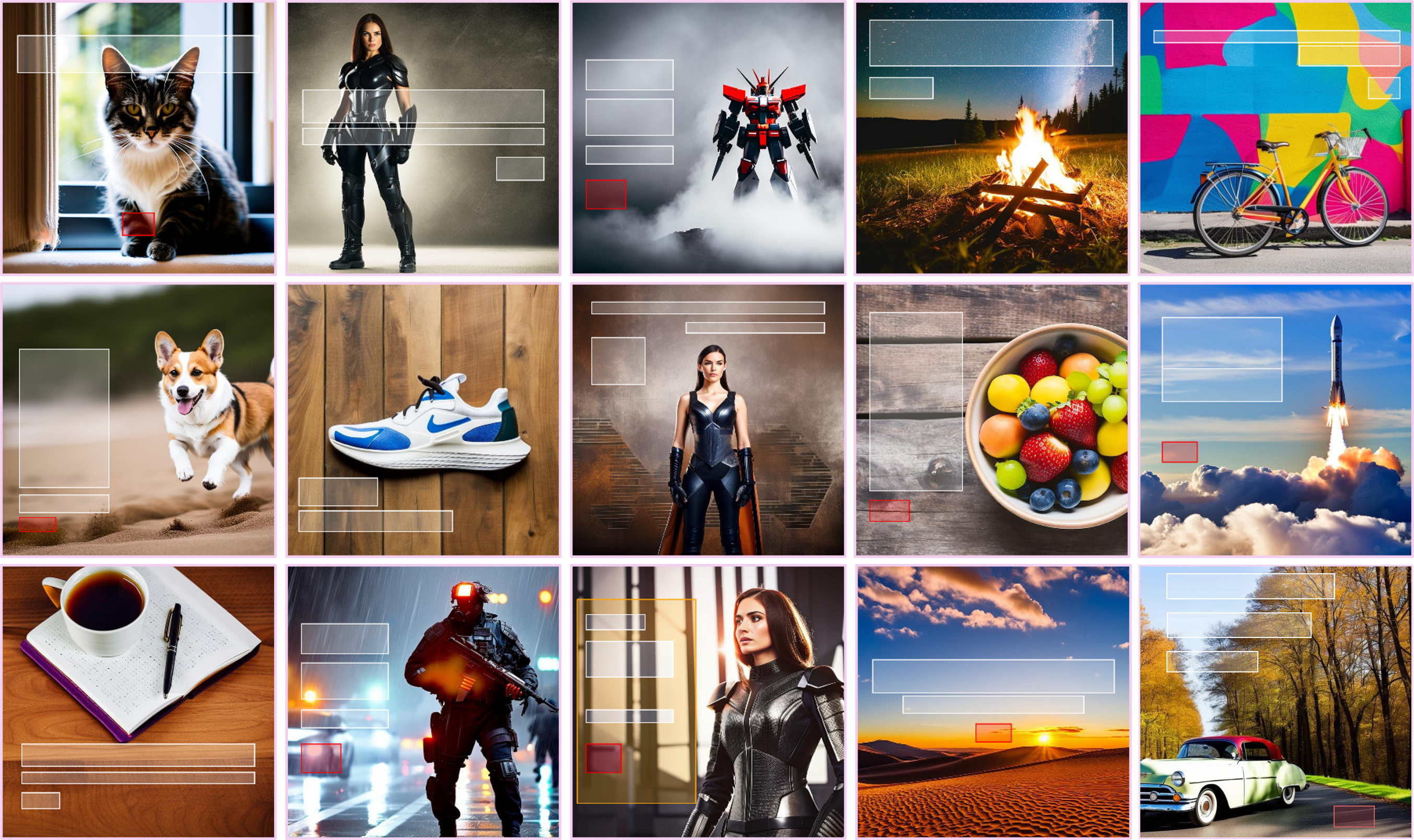}
\caption{
Additional results generated by our method.
White, orange, and red boxes denote text, underlay, and button, respectively.
}

\label{fig:more_res}
\end{figure*}

Figure~\ref{fig:qualitative_res} shows the visual comparison of design templates generated by different methods. Our model can generate high-quality, diverse layouts and visually pleasing background images. More importantly, due to its strong ability to capture the interaction between layout and background image, our model can produce harmonious layout-image compositions. In contrast, the results of OpenCOLE and Desigen suffer from various issues. Specifically, in the results of OpenCOLE, foreground elements are significantly misaligned and salient background areas are often occluded by foreground elements. The layouts of Desigen sometimes exhibit undesirable overlap between elements (column 4).
The background images generated by Desigen appear less visually appealing, mainly because its explicit occlusion avoidance bias could cause large empty regions (for foreground element placement) in generated background images (columns 3 and 5). Additional generated results by our method are shown in Figure~\ref{fig:more_res}.

\begin{table*}[!h]
\caption{
Ablation study on the communication module. For each metric except ones related to CLIP and FID, values closer to that of real data (bottom row) indicate better performance. The best results are in \textbf{bold}.
}
\centering
\resizebox{\linewidth}{!}{
\begin{tabular}{cccccccccccc}
\toprule
\multirow{2}{*}{Communication} & \multicolumn{3}{c}{Image} & \multicolumn{5}{c}{Layout} & \multicolumn{2}{c}{Template} \\
\cmidrule(lr){2-4} \cmidrule(lr){5-9} \cmidrule(lr){10-11}
& FID$\downarrow$ & CLIP$\uparrow$ & Saliency Ratio & LayoutFID$\downarrow$ & Align & Overlap & Occlusion & Readability & TemplateFID$\downarrow$ & TemplateCLIP$\uparrow$ \\
\midrule
{}None & \textbf{19.32} & 30.37 & 21.37\% & 0.16 & 0.38 & 12.87 & 30.74\% & 12.23\% & 99.14 & 3.23 \\
Layout $\rightarrow$ Image & 19.74 & 29.84 & 15.40\% & 0.17 & 0.35 & 13.20 & 28.40\% & 14.30\% & 97.45 & 3.22 \\
Image $\rightarrow$ Layout & 19.39 & \textbf{30.43} & 20.90\% & 0.16 & 0.34 & 13.60 & 27.80\% & 14.00\% & 96.20 & 3.24 \\
\rowcolor{tableblue}
Bidirectional (Ours) & 19.57 & 29.79 & \textbf{14.56\%} & \textbf{0.15} & \textbf{0.31} & \textbf{11.98} & \textbf{19.30}\% & \textbf{11.41\%} & \textbf{86.63} & \textbf{3.25} \\
\midrule
\textcolor{gray}{Real Data}
& \textcolor{gray}{--} & \textcolor{gray}{27.50} & \textcolor{gray}{14.17\%}
& \textcolor{gray}{--} & \textcolor{gray}{0.31} & \textcolor{gray}{9.34} & \textcolor{gray}{21.14\%} & \textcolor{gray}{8.53\%}
& \textcolor{gray}{--} & \textcolor{gray}{3.39} \\
\bottomrule
\end{tabular}
}
\label{tab:ablation_comm}
\end{table*}

\subsection{Analysis on the Communication Module}
\label{exp:comm}
\begin{figure*}[h]
    \centering
    \includegraphics[width=\textwidth]{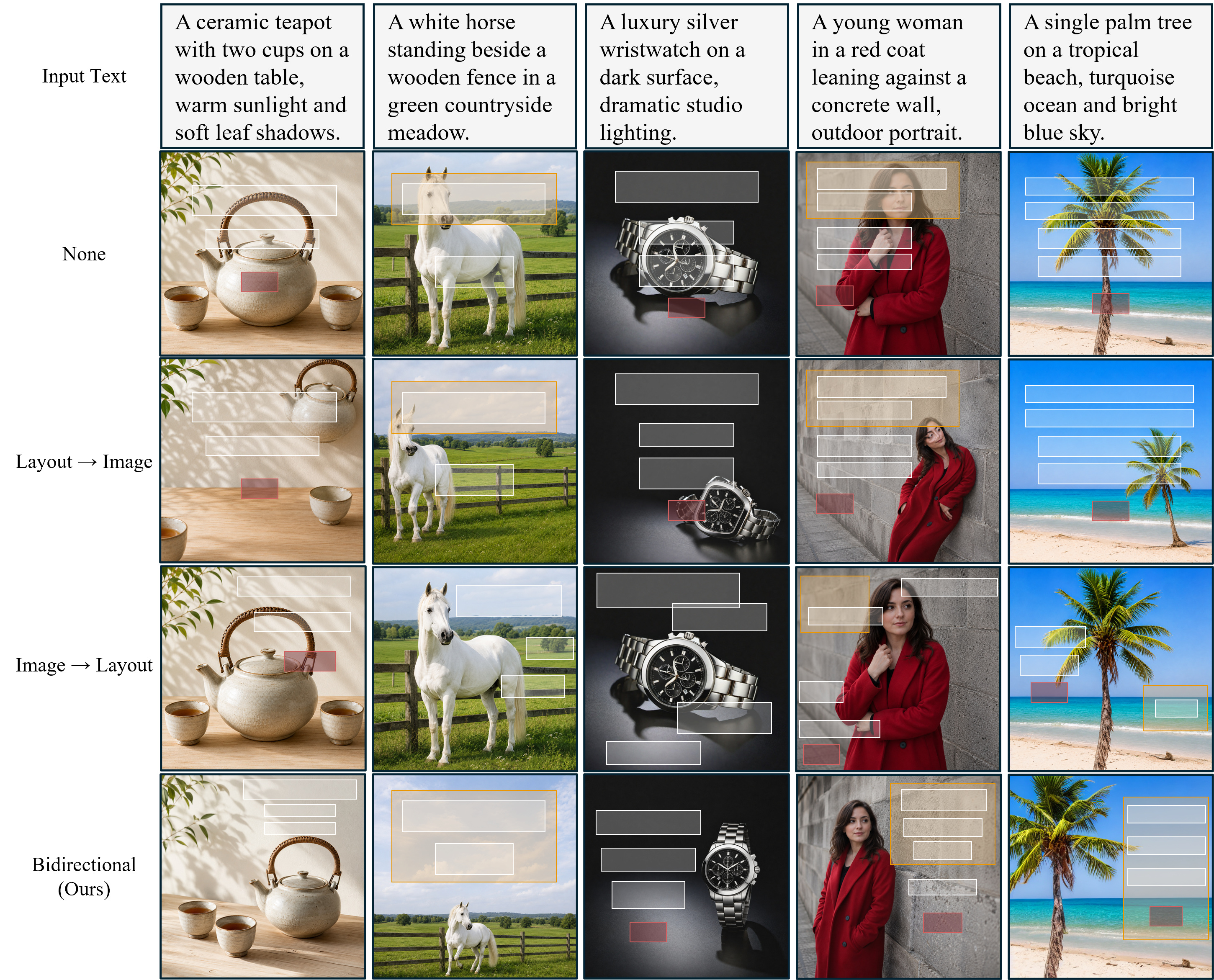}
    \caption{
    Qualitative comparison of our bidirectional communication module against its three ablated versions that allow no information exchange between image and layout (None), only allow layout-to-image information flow (Layout $\rightarrow$ Image), only allow image-to-layout information flow (Image $\rightarrow$ Layout). Each column uses the same input text and initial noise.
    }

    \label{fig:ablation_qualitative}
    \vspace{1.5em}
\end{figure*}

\noindent \textbf{Ablations.}
One key ingredient of our model is the communication module that enables the image and layout backbones to interact with each other during the generation process. To test the effect of this component, we consider a variant, where the communication module is disabled and thus the two backbones denoise \textit{independently}. As shown in Table~\ref{tab:ablation_comm},
using the communication module
dramatically reduces the saliency ratio, occlusion, and readability, and contributes to a significant improvement in TemplateFID. This highlights the importance of the communication module for improving the composition harmony of the background image and layout.
We further consider two variants that only allow unidirectional information flow from layout to image (by only augmenting the image representation $h^I$ using Equation \ref{eq:comm1})
and from image to layout (by only augmenting the layout representation $h^L$ using Equation \ref{eq:comm2}),
respectively. Compared with our communication module that enables \textit{bidirectional} information exchange, the unidirectional variants yield substantially worse performance on occlusion, readability, and TemplateFID. This suggests that bidirectional interaction between image and layout is critical for generating harmonious image-layout compositions. Figure~\ref{fig:ablation_qualitative} shows a visual comparison of results obtained using our communication module and the aforementioned variants.

\begin{figure*}[!h]
    \centering
    \subfloat{        \includegraphics[width=0.48\textwidth]{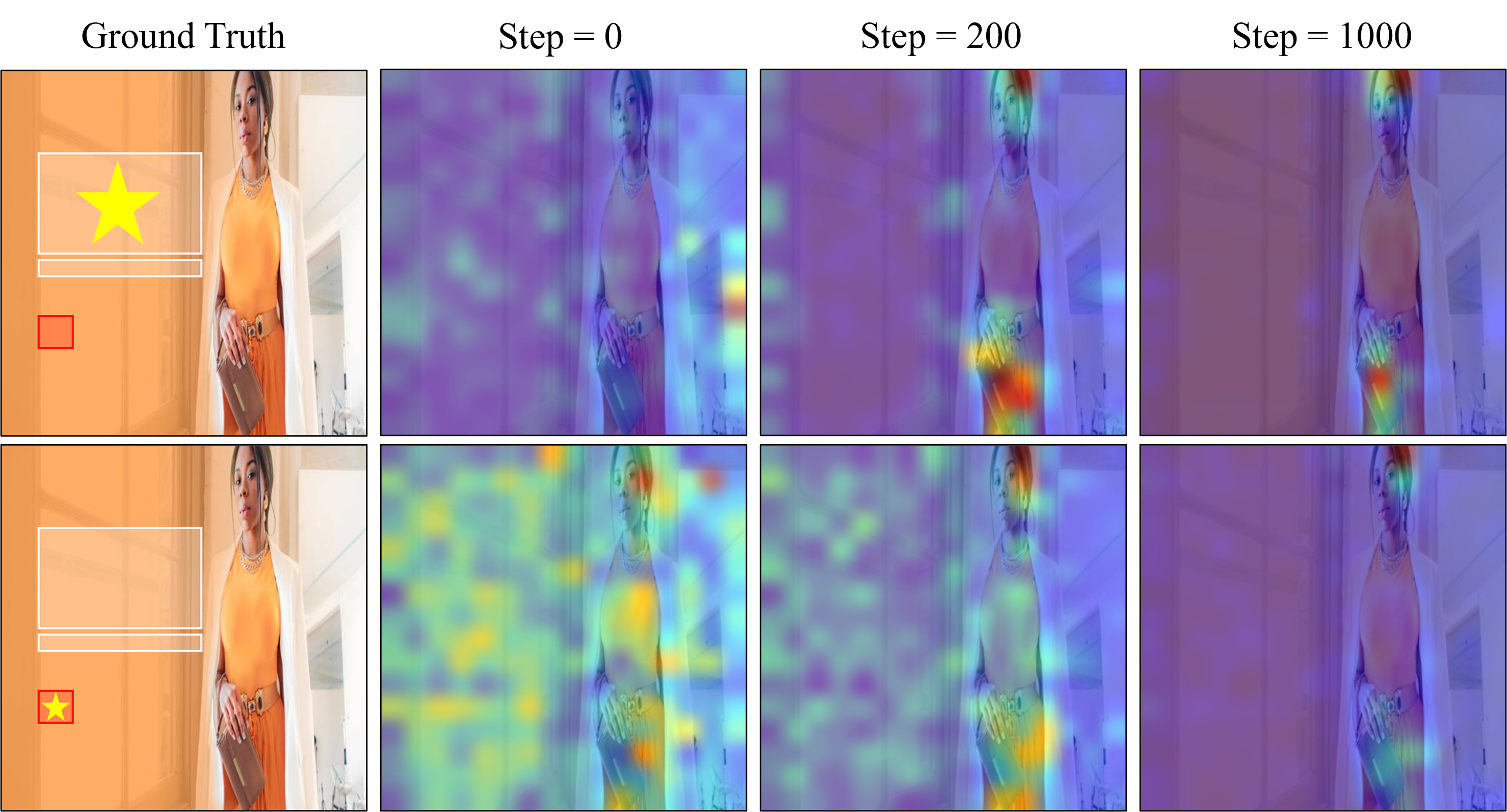}
        \label{fig:cross_attention_visualization}}
    \hfill
    \subfloat{        \includegraphics[width=0.48\textwidth]{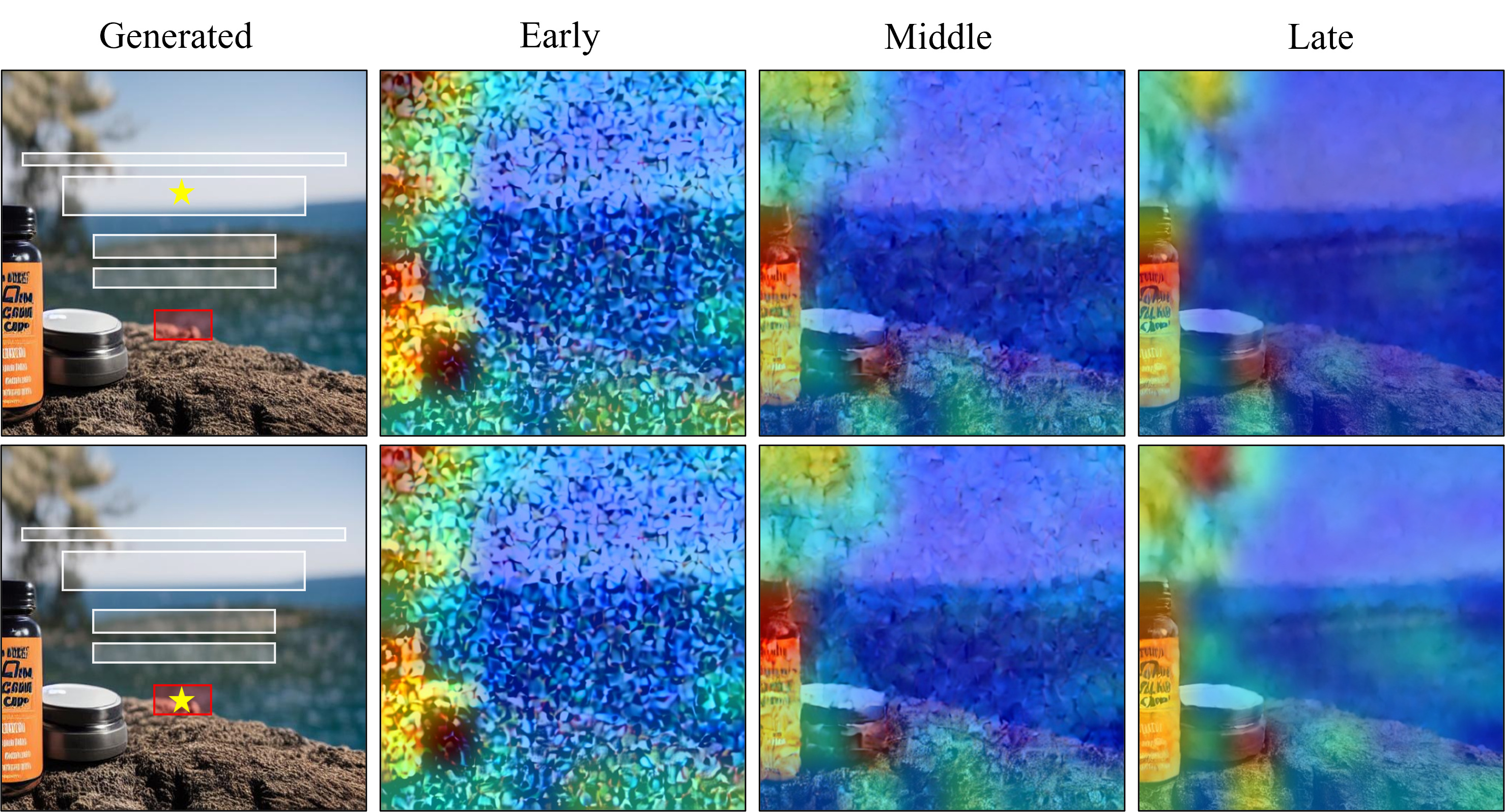}
        \label{fig:denoise_attention}}

    \caption{Layout-to-image cross-attention visualization in the communication module.
    \textit{Left}: During training, layout element tokens (marked by golden stars) progressively attend to salient regions as the number of training steps increases from 0 to 1000.
    \textit{Right}: At inference, they keep attending to salient image regions throughout all stages of the denoising process.
    }
    \label{fig:comm_attention}
\end{figure*}

\noindent \textbf{Visualization of Cross-Attention.}
To gain a more intuitive understanding of what the communication module learns, we visualize layout-to-image cross-attention during training and inference in Figure~\ref{fig:comm_attention}. During training, layout element tokens progressively focus on visually salient image regions, indicating that the module learns meaningful spatial relationships between layout elements and image content. At inference time, layout tokens already attend to salient regions at early denoising steps
and become increasingly concentrated as denoising proceeds.

\noindent \textbf{Communication Scheduling.}
In our implementation, we enable the communication module at the first $30 \%$ of the denoising process, which we find leads to overall good results. Let $\rho$ be the proportion of denoising steps in which the communication module is enabled; $\rho=100\%$ means the communication is enabled throughout all denoising steps, while $\rho=0\%$ disables the communication completely. We study the performance of our model at different $\rho$ values. The quantitative results are shown in Table~\ref{tab:comm_schedule}. \textit{No communication ($\rho=0\%$)} yields the lowest image FID and highest image CLIP, but at the cost of degraded image-layout harmony: it yields the highest TemplateFID and its occlusion deviates significantly from that of real data, indicating that it fails to achieve layouts that harmonize well with the backgrounds. \textit{Excessive communication ($\rho=100\%$)} leads to the highest layout FID and image FID, suggesting that excessive cross-modal information exchange harms layout and image quality. $\rho=30\%$ yields the best image-layout harmonization (the best results in TemplateFID, occlusion, readability), while achieving high layout quality (the near-optimal LayoutFID and the best alignment) and good image quality (competitive image FID and CLIP).

\begin{table*}[ht]
\caption{
Quantitative performance of our model by enabling the communication module at the first $\rho$ of denoising steps. For all metrics except CLIP and FID-related ones, values closer to those calculated from real data indicate better performance. Best and second-best are bold and underlined, respectively.
}
\centering
\resizebox{\linewidth}{!}{
\begin{tabular}{lcccccccccc}
\toprule
\multirow{2}{*}{$\rho$} & \multicolumn{3}{c}{Image} & \multicolumn{5}{c}{Layout} & \multicolumn{2}{c}{Template} \\
\cmidrule(lr){2-4} \cmidrule(lr){5-9} \cmidrule(lr){10-11}
& FID$\downarrow$ & CLIP$\uparrow$
& Saliency Ratio & LayoutFID$\downarrow$ & Align & Overlap & Occlusion & Readability & TemplateFID$\downarrow$ & TemplateCLIP$\uparrow$ \\
\midrule
$100\%$
& 22.84 & 27.97
& \textbf{14.31\%} & 0.51 & 0.41 & 15.18 & 18.97\% & 13.38\% & 97.95 & 3.20 \\

$70\%$
& 22.39 & 28.44
& 14.53\% & 0.36 & 0.37 & 14.41 & 18.94\% & 12.79\% & 95.92 & 3.21 \\

$50\%$
& 21.66 & 28.42
& \underline{14.34\%} & 0.24 & 0.35 & 13.19 & \underline{19.07\%} & 12.37\% & 91.45 & 3.22 \\

\rowcolor{tableblue}
$30\%$
& 19.57 & 29.79
& 14.56\% & \underline{0.15} & \textbf{0.31} & \underline{11.98} & \textbf{19.30\%} & \textbf{11.41\%} & \textbf{86.63} & \textbf{3.25} \\

$10\%$
& \underline{19.45} & \underline{29.87}
& 19.07\% & \textbf{0.14} & \underline{0.32} & \textbf{11.41} & 26.76\% & 13.27\% & \underline{89.42} & 3.22 \\

$0\%$
& \textbf{19.32} & \textbf{30.37}
& 21.37\% & 0.16 & 0.38 & 12.87 & 30.74\% & \underline{12.23\%} & 99.14 & \underline{3.23} \\

\midrule
\textcolor{gray}{Real Data}
& \textcolor{gray}{--} & \textcolor{gray}{27.5} & \textcolor{gray}{14.17\%}
& \textcolor{gray}{--} & \textcolor{gray}{0.31} & \textcolor{gray}{9.34} & \textcolor{gray}{21.14\%} & \textcolor{gray}{8.53\%} & -- & 3.39 \\

\bottomrule
\end{tabular}
}
\label{tab:comm_schedule}
\end{table*}

\begin{figure*}[ht]
    \centering
    \includegraphics[width=\textwidth]{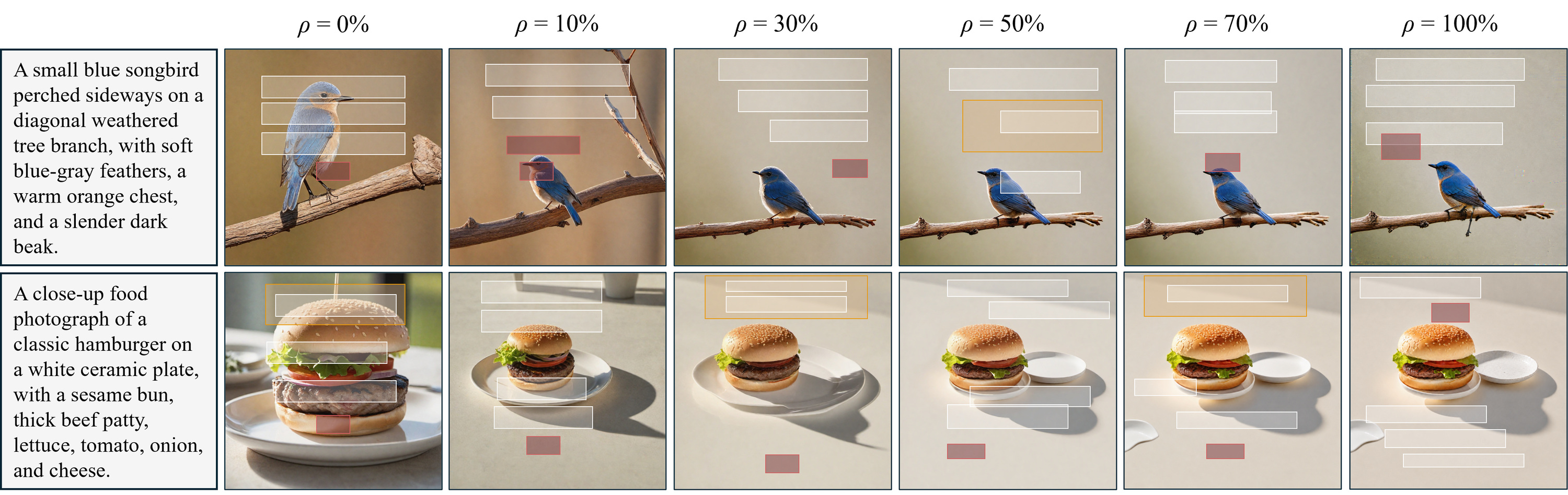}
    \caption{
    Qualitative results of enabling the communication module at the first $\rho$ of the denoising steps. For each input text, all the results use identical initial noise and only $\rho$ is varied. Our implementation uses $\rho = 30 \%$.
    }

    \label{fig:rho_qualitative}
\end{figure*}

\begin{table*}[htbp]
\caption{
Quantitative results of applying occlusion-aware guidance (OAG) and readability-aware guidance (RAG) to our model (Ours).
For each metric except ones related to CLIP and FID, values closer to that of real data (bottom row) indicate better performance. The best and second-best numerical results are highlighted in \textbf{bold} and \underline{underlined}, respectively.
}
\centering
\begin{adjustbox}{width=\textwidth}
\begin{tabular}{lcccccccccc}
\toprule
\multirow{2}{*}{Method} & \multicolumn{3}{c}{Image} & \multicolumn{5}{c}{Layout} & \multicolumn{2}{c}{Template} \\
\cmidrule(lr){2-4} \cmidrule(lr){5-9} \cmidrule(lr){10-11}
& FID$\downarrow$ & CLIP$\uparrow$ & Saliency Ratio
& LayoutFID$\downarrow$ & Align & Overlap & Occlusion & Readability
& TemplateFID$\downarrow$ & TemplateCLIP$\uparrow$  \\
\midrule
Ours
& \textbf{19.57} & \textbf{29.79} & \textbf{14.56\%}
& \textbf{0.15} & \textbf{0.31} & \textbf{11.98} & \textbf{19.30\%} & 11.41\%
& \textbf{86.63} & \textbf{3.25} \\

\addlinespace[0.1pt]
\cmidrule(lr){1-11}
\addlinespace[0.1pt]

Ours + OAG
& \textbf{19.57} & \textbf{29.79} & \textbf{14.56\%}
& \underline{0.21} & \underline{0.32} & \underline{13.26} & 12.66\% & \underline{10.35\%}
& \underline{92.81} & \underline{3.23} \\

Ours + RAG
& \textbf{19.57} & \textbf{29.79} & \textbf{14.56\%}
& 0.24 & 0.34 & 14.13 & 12.46\% & \textbf{7.49}\%
& 96.78 & 3.20 \\

\midrule
\textcolor{gray}{Real Data}
& \textcolor{gray}{--} & \textcolor{gray}{27.50} & \textcolor{gray}{14.17\%}
& \textcolor{gray}{--} & \textcolor{gray}{0.31} & \textcolor{gray}{9.34} & \textcolor{gray}{21.14\%} & \textcolor{gray}{8.53\%}
& \textcolor{gray}{--} & \textcolor{gray}{3.39} \\
\bottomrule
\end{tabular}
\end{adjustbox}
\label{tab:guidance}
\end{table*}

\begin{figure*}[t]
\centering
\includegraphics[width=\textwidth]{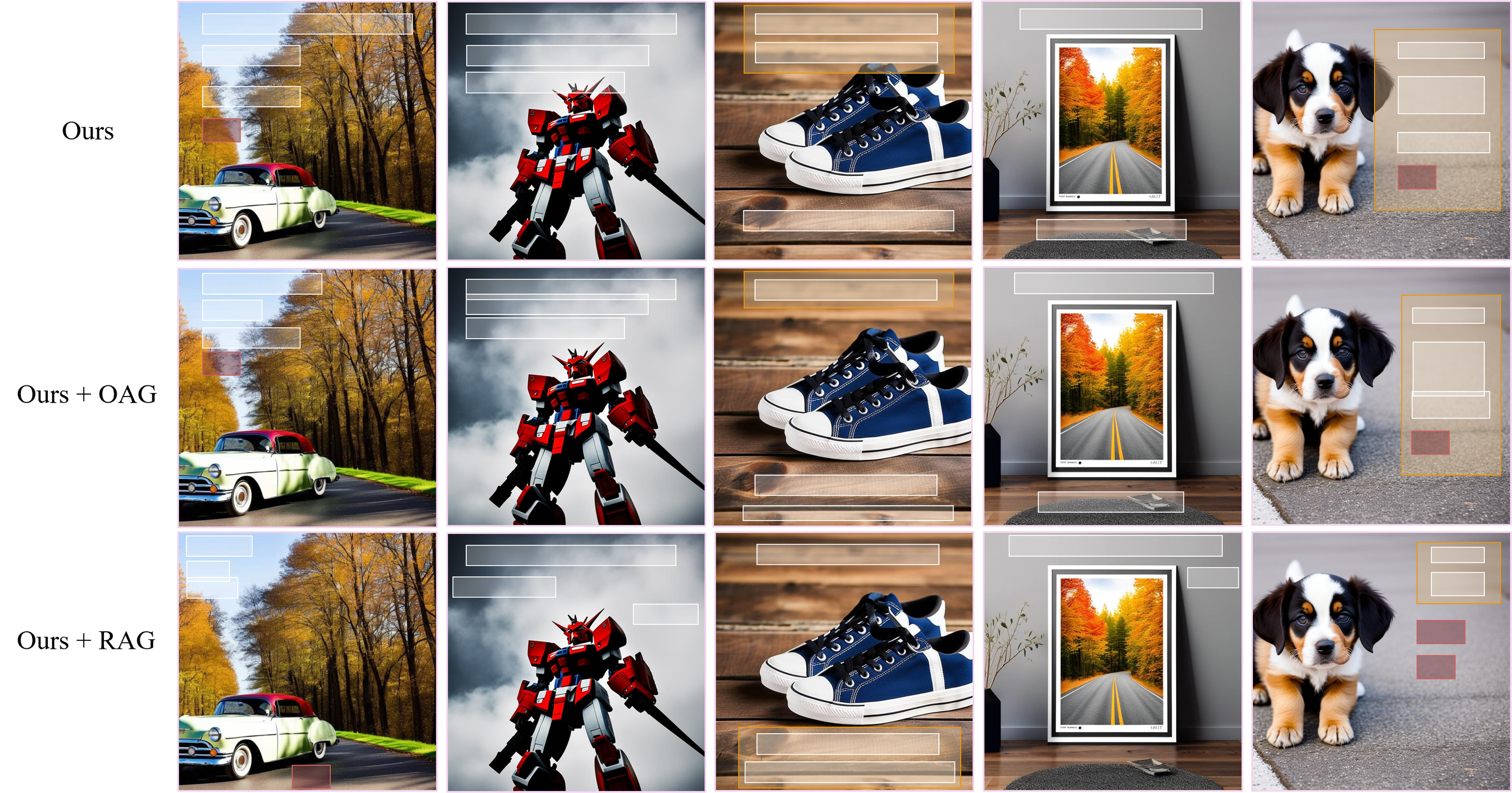}
\caption{
Qualitative results of preference-based guidance. Occlusion-aware guidance (OAG) prevents foreground elements from occluding important background regions; readability-aware guidance (RAG) improves text readability by putting foreground elements over flat background regions.
White, orange, and red boxes denote text, underlay, and button, respectively.
}

\label{fig:guidance}
\end{figure*}

\subsection{Preference-based Guidance}
As shown in Table~\ref{tab:guidance}, applying occlusion-aware guidance (OAG) results in a significant reduction in occlusion from $19.30\%$ to $12.66\%$, which means that occlusion of salient background regions is dramatically reduced. Using readability-aware guidance improves readability
by a large margin (from $11.41\%$ to $7.49\%$), suggesting that it effectively improves text readability. Note that enabling preference-based guidance does compromise image quality, while leading to only minor degradation in other metrics (LayoutFID, alignment, overlap, TemplateFID, and TemplateCLIP). Figure~\ref{fig:guidance} visually compares the results of our model against the variants obtained by applying OAG and RAG. It can be seen that OAG prevents foreground layout elements from occluding visually important objects in the background, while RAG prefers to place foreground elements over uniform background regions.

\section{Conclusion}
\label{sec:conclusion}

In this paper, we present a generative model for graphic design templates based on a joint image-layout generation paradigm that contrasts with the sequential generation scheme commonly used in prior work. Our model is constructed within a latent diffusion framework and uses a specialized communication module to explicitly learn image-layout interaction and jointly generate both an image and a layout in a single generative process. Our model is able to faithfully capture the joint distribution of images and layouts, thereby achieving high image-layout harmony. We also introduce a training-free guidance technique, which allows users to shift generated results away from the training data distribution and move them towards their preferred design patterns.
Our experimental results demonstrate that our model is superior to existing methods, synthesizing visually appealing background images, high-quality layouts, and harmonious image-layout compositions.

\makeatletter
\setlength{\@fptop}{0pt}
\setlength{\@fpsep}{4pt}
\setlength{\@fpbot}{0pt plus 1fil}
\makeatother
\setlength{\floatsep}{6pt}
\setlength{\textfloatsep}{6pt}
\setlength{\abovecaptionskip}{2pt}
\setlength{\belowcaptionskip}{0pt}

\bibliographystyle{IEEEtran}
\bibliography{main}

\clearpage
\appendices
\setcounter{figure}{0}
\setcounter{table}{0}
\setcounter{equation}{0}
\setcounter{promptbox}{0}
\renewcommand{\thefigure}{S\arabic{figure}}
\renewcommand{\thetable}{S\arabic{table}}
\renewcommand{\theequation}{S\arabic{equation}}
\renewcommand{\thepromptbox}{S\arabic{promptbox}}
\markboth{IEEE Transactions on Visualization and Computer Graphics}{Yang \MakeLowercase{\textit{et al.}}: InterIL Supplementary Material}
\section{Supplementary Material}

\subsection{Interpretability and Effectiveness of the Communication Module}
\label{supp:sec:cross_modal_attention}

\subsubsection{Training Stage Attention Map Visualization}
To better understand how the layout and image modalities interact during training,
we visualize cross-attention maps in both directions: from layout tokens to image patches
and from image patches to layout tokens.

\begin{figure*}[htbp]
    \centering
    \includegraphics[width=\linewidth]{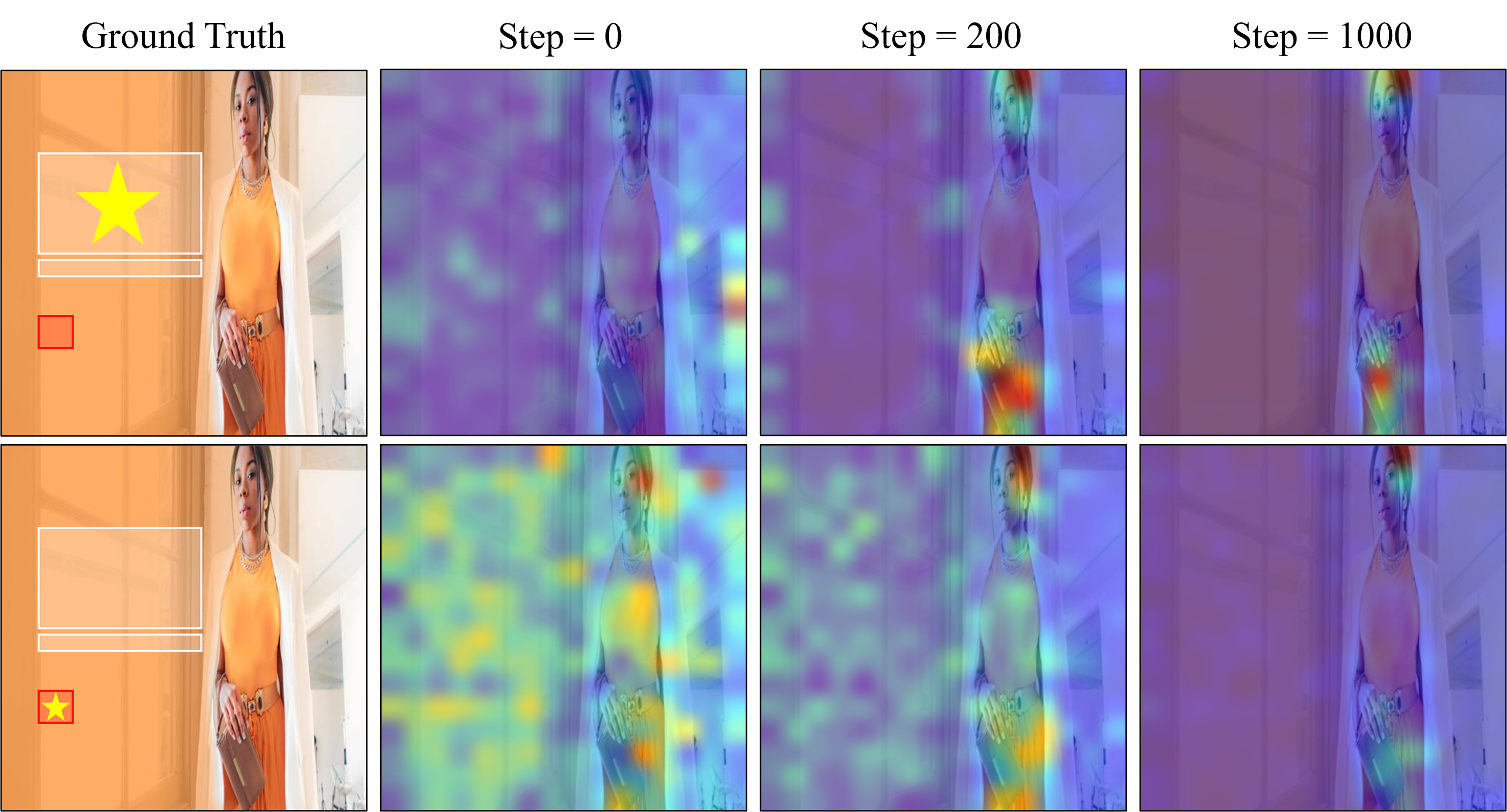}
    \caption{
        Visualization of layout-to-image cross-attention across training.
        The \textbf{leftmost} column shows the ground-truth layout, where the bounding box
        marked with a \textbf{golden star} indicates the selected layout token.
        The three images to the right illustrate how this token attends to different regions
        of the background image at training steps 0, 200, and 1000.
        As training progresses, the attention becomes increasingly focused on salient objects,
        indicating that the communication module gradually learns meaningful spatial correspondences
        between layout elements and image content.}

    \label{supp:fig:cross_attention_visualization}
\end{figure*}

As shown in Figure~\ref{supp:fig:cross_attention_visualization},
the attention from layout tokens to image features evolves significantly during training.
In the early stage (training step 0), each layout token attends almost uniformly to the entire image,
without clear spatial preference.
As training progresses, however, these attentions become sharper and more semantically meaningful:
after 1000 steps, layout tokens are able to consistently focus on salient objects and object boundaries
(e.g., text regions or buttons), demonstrating that the communication module learns to
capture the spatial layout of the background image and align layout elements with visual context.

\begin{figure*}[htbp]
    \centering
    \includegraphics[width=0.9\linewidth]{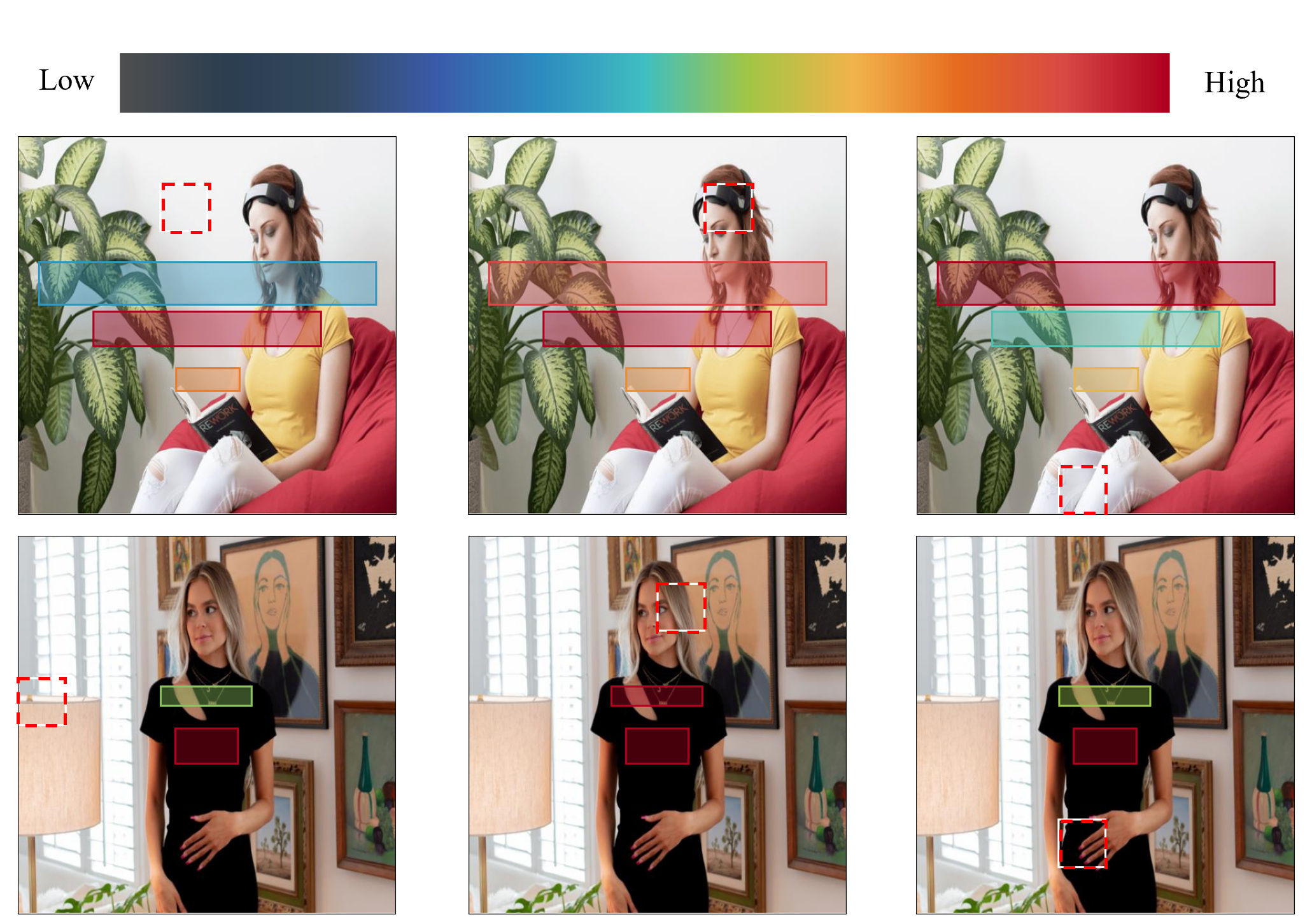}
    \caption{
        Cross-attention visualization from image patches to layout elements at different denoising timesteps. Each red dashed box marks a selected image patch, and the color overlay on each layout element (red = high attention, blue = low attention) indicates how strongly that patch influences the layout token. We observe that patches located in \textbf{visually salient regions} (e.g., faces, objects) exert much stronger influence on nearly all layout boxes, whereas patches from \textbf{non-salient background areas} contribute weakly and only to a subset of elements.}

    \label{supp:fig:img2layout}
\end{figure*}

Complementary to this observation, Figure~\ref{supp:fig:img2layout} presents the reverse attention flow,
from image patches to layout tokens. We find that patches corresponding to \textbf{highly salient regions}
(e.g., human faces) have a much stronger influence on the layout: their attention maps exhibit consistently
high responses across almost all bounding boxes. In contrast, patches from \textbf{less salient background areas}
tend to affect only a subset of layout tokens and show much weaker activations overall.

\subsubsection{Inference Stage Attention Map Visualization}
\begin{figure*}[t]
  \centering
  \includegraphics[width=\textwidth]{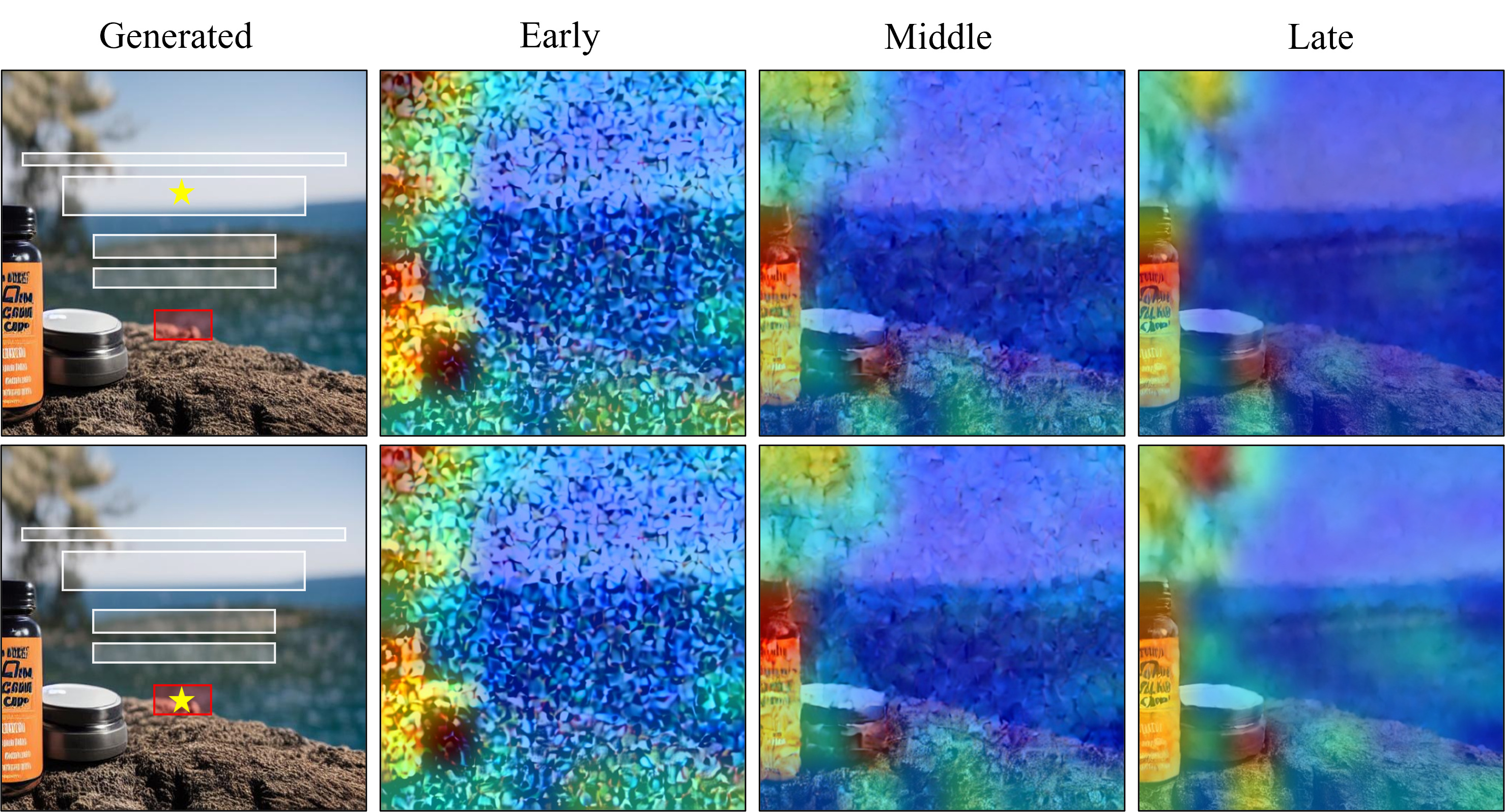}
  \caption{
    Cross-attention visualization from layout tokens to image patches across
    early, middle, and late denoising stages. For several representative layout
    tokens, we highlight the image regions they attend to. Even at early steps,
    the tokens already focus on salient objects, and this focus becomes sharper
    as denoising progresses, indicating that the communication module learns
    meaningful cross-modal correspondences throughout the generation process.
    }

  \label{supp:fig:denoise_attention}
\end{figure*}

We further inspect the cross-attention weights from layout tokens to image patches during inference
as the denoising process unfolds. As shown in Figure~\ref{supp:fig:denoise_attention},
even at early timesteps ($t=701$), layout tokens already focus on salient regions of the image,
such as objects and prominent anchors. As denoising progresses, these attentions sharpen and stabilize,
indicating that the communication module dynamically refines its cross-modal grounding.
Importantly, this behavior is consistent with the training stage observations:
the communication module preserves the spatial alignment ability it learned during training
and can readily attend to salient regions and boundaries in the image during inference.
This consistency suggests that the learned cross-modal alignment is not only acquired in training,
but also effectively retained and exploited during inference.

\subsection{GPT-Based Evaluation}
\label{supp:sec:gpt_based_eval}

\subsubsection{GPT-5 Evaluation}
\label{supp:sec:gpt4v_eval}
Inspired by recent work on graphic design evaluation~\cite{jia2023cole,inoue2024opencole},
we leverage GPT-5 to automatically assess the quality of generated design templates.
Given a design template, GPT-5 is instructed to evaluate it across four complementary aspects,
each scored on a scale of 1 to 10, where higher values indicate better performance.
For each candidate model (Desigen, OpenCOLE, and ours), we generate 2,000 design templates
using text inputs randomly sampled from the test set.
The aggregated GPT-5 scores are reported in Table~\ref{supp:tab:gpt4v_eval}.
To ensure consistency, we design a specialized prompt tailored for template evaluation,
as detailed in PromptBox~\ref{supp:box:gpt4v_prompt}.

The four evaluation aspects capture different dimensions of design quality:

\begin{itemize}[leftmargin=*]
    \item \textit{Image Quality:}
    Measures the visual fidelity, realism, and aesthetic appeal of the generated background image.

    \item \textit{Layout Quality:}
    Evaluates the structural integrity and usability of the layout, including alignment, spacing,
    proportions, overlap avoidance, and readability.

    \item \textit{Image--Layout Harmony:}
    Assesses how well the generated layout integrates with the background image, including visibility,
    contrast, and avoidance of distracting conflicts with salient regions.

    \item \textit{Text--Design Relevance:}
    Evaluates whether the generated template is semantically relevant to the input text prompt
    in both visual content and overall design intent.
\end{itemize}

As shown in Table~\ref{supp:tab:gpt4v_eval}, our model consistently outperforms existing baselines across all four evaluation aspects. In Image Quality, it achieves 6.94, higher than Desigen (6.42) and OpenCOLE (6.81), and close to real data (7.44). In Layout Quality, our method reaches 7.76, substantially improving over Desigen (6.54) and OpenCOLE (6.02), and approaching real data (7.87). The advantage is also clear in Image--Layout Harmony, where our score of 8.07 exceeds Desigen (7.47) and OpenCOLE (5.84), indicating more coherent multimodal compositions. Finally, in Text--Design Relevance, our score of 7.34 is notably better than Desigen (6.21) and OpenCOLE (6.38), and close to real data (7.49), suggesting stronger semantic alignment between the generated template and the input prompt. Overall, these GPT-5 results provide additional evidence that our method produces more coherent and relevant design templates than both Desigen and OpenCOLE.

\begin{table}[htbp]
\caption{
GPT-5 evaluation results across four aspects of design quality:
(i) Image Quality,
(ii) Layout Quality,
(iii) Image--Layout Harmony,
(iv) Text--Design Relevance.
The best and second-best results are shown in \textbf{bold} and \underline{underlined}, respectively.
}
\label{supp:tab:gpt4v_eval}
\centering
\scriptsize
\setlength{\tabcolsep}{3pt}
\begin{tabular}{lcccc}
\toprule
Method & (i) & (ii) & (iii) & (iv) \\
\midrule
Desigen
& 6.42 & \underline{6.54} & \underline{7.47} & 6.21 \\

OpenCOLE
& \underline{6.81} & 6.02 & 5.84 & \underline{6.38} \\

Ours
& \textbf{6.94} & \textbf{7.76} & \textbf{8.07} & \textbf{7.34} \\
\midrule
\textcolor{gray}{Real Data}
& \textcolor{gray}{7.44} & \textcolor{gray}{7.87} & \textcolor{gray}{8.39} & \textcolor{gray}{7.49} \\
\bottomrule
\end{tabular}
\end{table}

\refstepcounter{promptbox}
\begin{figure*}[!t]
\label{supp:box:gpt4v_prompt}
\begin{tcolorbox}[mypromptbox,title={PromptBox~\thepromptbox: System Prompt for GPT-5 Evaluation}]
\small
\begin{multicols}{2}
\raggedcolumns

You are an autonomous AI Assistant who evaluates graphic design templates with objectivity, precision, and consistency.

Your goals are: Provide unbiased and actionable critiques of design templates based on established visual design principles. Assess image quality, layout quality, image-layout harmony, and text-design relevance. Maintain strict scoring standards and concise reasoning.

All coordinates follow the convention that the upper-left corner is the origin, the x-axis increases rightward, and the y-axis increases downward.

Please abide by the following rules: Score as objectively as possible. A flawless template may score 10 points, a mediocre one around 7, a template with clear issues around 4, and a very poor template 1--2 points. Keep reasoning brief.

\textbf{Grading criteria:}

\begin{itemize}[leftmargin=*]
    \item \textbf{Image Quality (1--10):}
    Evaluate the visual fidelity, realism, and aesthetic appeal of the background image.
    High scores reflect sharp details, natural textures, clean rendering, and an overall pleasing composition.
    Low scores indicate blurriness, artifacts, noise, distortions, or visually incoherent backgrounds.

\end{itemize}
\columnbreak
\begin{itemize}[leftmargin=*]

    \item \textbf{Layout Quality (1--10):}
    Evaluate the intrinsic structural soundness and usability of the bounding-box layout.
    High-scoring layouts exhibit precise alignment, consistent spacing, balanced proportions, minimal overlap,
    low unintended occlusion, and clear readability.
    Low scores reflect misalignment, irregular spacing, excessive overlap, poorly sized regions,
    or other structural flaws that hinder clarity and usability.

    \item \textbf{Image--Layout Harmony (1--10):}
    Evaluate how effectively the layout interacts with the background image.
    High scores indicate strong contrast, low visual interference, minimal conflict with salient regions,
    and placements that enhance visibility and focus.
    Low scores reflect distracting overlaps, poor visibility, or layout elements placed on cluttered
    or visually dominant areas that reduce clarity.

    \item \textbf{Text--Design Relevance (1--10):}
    Evaluate whether the generated design is semantically relevant to the input text prompt.
    High scores indicate that the visual content, layout decisions, and overall style match the intended theme,
    product, or message expressed in the prompt.
    Low scores indicate weak semantic correspondence, mismatched imagery, or a design that fails to reflect
    the intended content or communicative purpose.
\end{itemize}

\end{multicols}
\end{tcolorbox}
\end{figure*}

\subsection{Blind Human Preference Study on Generated Templates}
\label{supp:sec:human_preference_study}

To complement the automatic and metric-based evaluations, we conduct a blind forced-choice study comparing templates generated by InterIL, Desigen, and OpenCOLE. The study involves 32 participants and 40 text prompts. For each prompt, outputs from the three methods are presented with their method identities hidden and their display order randomized. Participants are asked to select their preferred result separately according to four aspects: image quality, layout quality, image--layout harmony, and text--design relevance.

Across all responses, InterIL is selected as the preferred result in 68\% of comparisons for image quality, 78\% for layout quality, 83\% for image--layout harmony, and 71\% for text--design relevance. These human judgments are consistent with the GPT-5 results in Table~\ref{supp:tab:gpt4v_eval} and with the TemplateFID ranking, providing direct perceptual evidence for the quality and coherence of the generated templates.

\subsection{Preference-Specific Guidance}
\label{supp:sec:occ_guidance}

\subsubsection{Occlusion-Aware Guidance (OAG)}
\begin{figure*}[htbp]
    \centering
    \includegraphics[width=\linewidth]{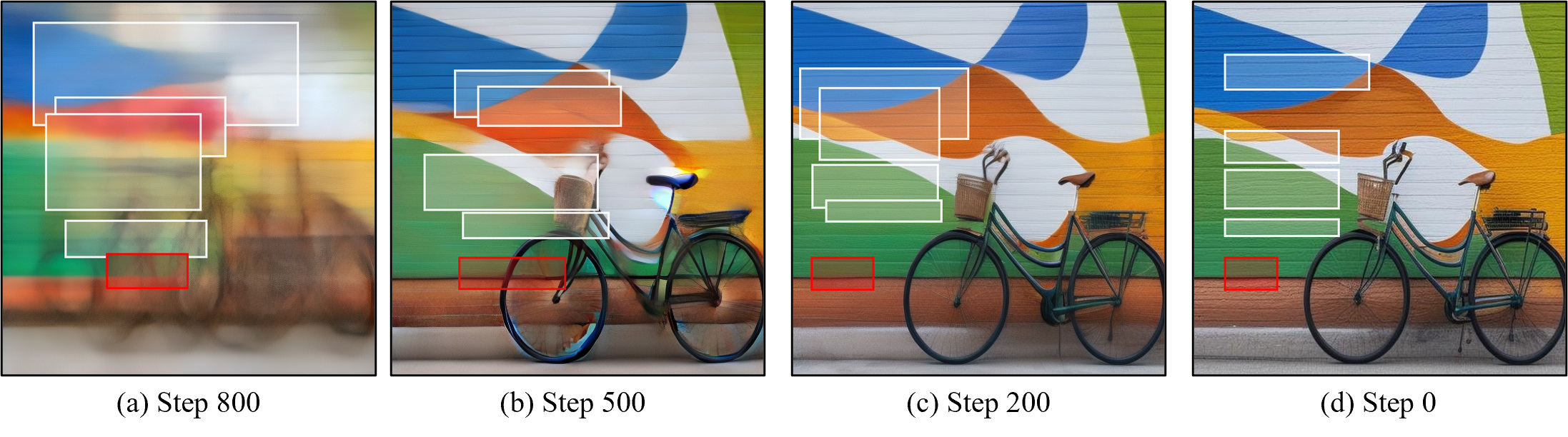}
    \caption{
        Denoising trajectory under occlusion-aware guidance (\textit{occ scale} = 3),
        shown as the first guidance example in our framework.
        The layout is gradually pushed away from salient image regions as denoising progresses,
        demonstrating how OAG steers the model toward low-occlusion configurations.
    }

    \label{supp:fig:occ_steps}
\end{figure*}

To illustrate how the proposed inference-time guidance framework can accommodate
different user preferences, we first analyze occlusion-aware guidance (OAG) as one
concrete example. To qualitatively assess its effect during generation, we visualize
the denoising trajectory of a sample under guidance scale $s=3$.
As shown in Figure~\ref{supp:fig:occ_steps}, the layout elements initially overlap
with salient regions of the image (e.g., foreground objects).
As denoising proceeds, however, these elements are gradually pushed away from
the salient content and repositioned into less intrusive areas.
This stepwise visualization demonstrates how one guidance objective can steer the
layout toward a user-preferred low-occlusion solution at test time.

\subsubsection{Ablation on OAG Scale}
We conduct an ablation study on the OAG guidance scale $s$ to better understand
this particular guidance instantiation. Since OAG directly influences only the
placement of layout elements, we evaluate both layout-level and template-level
metrics (Table~\ref{supp:tab:occ_ablation_layout}).

As $s$ increases, the model becomes more conservative in placing layout elements over salient regions.
Occlusion decreases consistently and reaches its minimum at larger scales (e.g., $s=10$),
indicating that stronger OAG more aggressively enforces avoidance of important background areas.
Readability also improves with moderate scales (around $s=3$ to $s=5$),
but excessive values lead to diminishing returns.
At the same time, stronger guidance introduces side effects:
LayoutFID and alignment gradually deteriorate, and template-level realism is harmed.
Specifically, TemplateFID rises steadily (e.g., 86.63 $\to$ 114.34),
while TemplateCLIP drops slightly (3.25 $\to$ 3.16),
suggesting that overly strong preference steering distorts the natural distribution of layouts
and weakens overall design quality.

\begin{figure*}[htbp]
    \centering
    \includegraphics[width=0.85\linewidth]{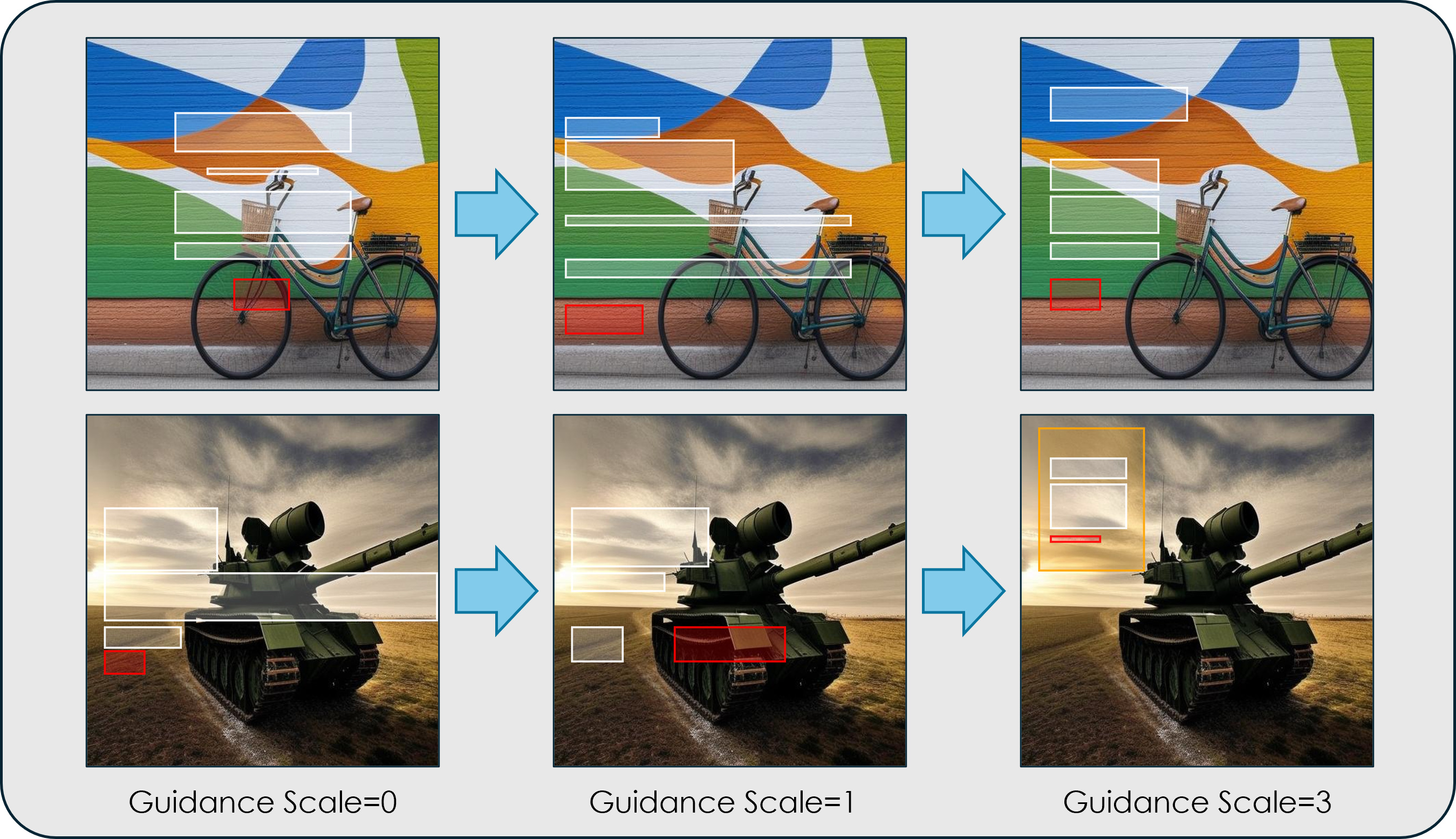}
    \caption{
        Visualization of two samples under different OAG scales.
        From left to right: guidance scale = 0, 1, and 3.
        The comparison shows how stronger OAG encourages layout elements
        to avoid salient regions more consistently, resulting in cleaner and more readable templates.
    }

    \label{supp:fig:occ_guidance_and_steps}
\end{figure*}

A moderate guidance scale achieves the best trade-off.
In particular, $s=2$ and $s=3$ strike a balance between reducing occlusion and
maintaining competitive LayoutFID, alignment, and template-level realism.
Qualitative results in Figure~\ref{supp:fig:occ_guidance_and_steps} confirm this trend:
at $s=3$, layout elements are already well separated from salient objects in the background,
resulting in cleaner and more harmonious compositions without sacrificing fidelity.
Therefore, we adopt $s=3$ as the default OAG setting in our main experiments.
Larger scales (e.g., $s=7$ or $s=9$) can be useful when strict occlusion avoidance is preferred,
but they may come at the cost of reduced layout fidelity and visual realism.
This trade-off is central to the overall paper narrative: the unguided model best matches the
joint data distribution, whereas guidance is intended to provide additional flexibility for users
with more specific preferences.

\begin{table*}[htbp]
\centering
\caption{Ablation on different OAG scales ($s$) using 50 DDIM steps. This table analyzes one concrete guidance instantiation rather than the core unguided model itself. For each metric, the best result is marked in \textbf{bold}, and the second best is \underline{underlined}. The last row shows statistics computed from real data as reference.}
\resizebox{\linewidth}{!}{
\begin{tabular}{lccccccc}
\toprule
\multirow{2}{*}{Setting} & \multicolumn{5}{c}{Layout} & \multicolumn{2}{c}{Template} \\
\cmidrule(lr){2-6} \cmidrule(lr){7-8}
& L\mbox{-}FID$\downarrow$ & Align & Overlap & Occlusion & Readability & TemplateFID$\downarrow$ & TemplateCLIP$\uparrow$ \\
\midrule
s=0  & \textbf{0.148} & \textbf{0.31} & 11.98 & \textbf{19.30\%} & 11.41\% & 86.63 & 3.25 \\
s=1  & \underline{0.1547} & 0.42 & 11.90 & \underline{15.40\%} & 10.50\% & 87.05 & 3.25 \\
s=2  & 0.1874 & \underline{0.38} & \textbf{11.38} & 13.40\% & 10.27\% & 89.41 & 3.24 \\
s=3  & 0.2113 & 0.32 & 13.26 & 12.66\% & 10.35\% & 92.81 & 3.23 \\
s=4  & 0.2274 & 0.36 & \underline{11.80} & 12.21\% & 10.25\% & 94.35 & 3.23 \\
s=5  & 0.2899 & 0.34 & 13.47 & 11.22\% & 10.26\% & 104.83 & 3.22 \\
s=6  & 0.2946 & 0.41 & 13.06 & 11.22\% & 10.19\% & 104.32 & 3.21 \\
s=7  & 0.3324 & 0.34 & 13.79 & 10.35\% & \textbf{9.97\%} & 109.44 & 3.21 \\
s=8  & 0.3658 & 0.48 & 13.58 & 10.07\% & \underline{10.01\%} & 112.37 & 3.19 \\
s=9  & 0.3689 & 0.46 & 13.85 & 10.25\% & 10.22\% & 112.48 & 3.18 \\
s=10 & 0.4426 & 0.47 & 13.73 & 9.71\% & 10.05\% & 114.34 & 3.16 \\
\midrule
\textcolor{gray}{Real Data}
& \textcolor{gray}{--} & \textcolor{gray}{0.31} & \textcolor{gray}{9.34} & \textcolor{gray}{21.14\%} & \textcolor{gray}{8.53\%} & \textcolor{gray}{--} & \textcolor{gray}{3.39} \\
\bottomrule
\end{tabular}
}
\label{supp:tab:occ_ablation_layout}
\end{table*}

\subsubsection{Readability-Aware Guidance (RAG)}
RAG is the second guidance example considered in the paper.
It uses the same inference-time steering mechanism as OAG, but replaces the
saliency-based objective with a clutter-based readability objective defined on
text regions. This variant is useful when users specifically prefer text to be
placed on smoother, lower-texture background regions. Consistent with the
quantitative results in the main paper, RAG improves readability-oriented
preference metrics, but it may also reduce holistic realism and fidelity to the
joint data distribution. As with OAG, RAG should therefore be interpreted as an
optional preference-specific control rather than a replacement for the base model.

\subsection{User Study on Template Embedding Similarity}
\label{supp:sec:user_study}

To further validate the perceptual quality of our learned template embedding space,
we conducted a user study involving 104 participants, including 58 design experts
and 46 non-experts. We include non-experts because design templates are also consumed
and adapted by general users in practical applications. Each participant was shown triplets of the form
$(x, \tilde{x}_{\text{rand}}, \tilde{x}_{\text{emb}})$, where $x$ is a reference template,
$\tilde{x}_{\text{rand}}$ is a randomly selected template from the Web-design dataset,
and $\tilde{x}_{\text{emb}}$ is the most similar template to $x$ retrieved by our
learned embeddings. Although the embedding model is trained on a composed dataset that
excludes Web-design, participants were asked to choose, based on their intuition,
which of $\tilde{x}_{\text{rand}}$ or $\tilde{x}_{\text{emb}}$ is more similar to $x$,
considering both layout and image content.

\begin{figure*}[htbp]
    \centering
    \includegraphics[width=0.9\linewidth]{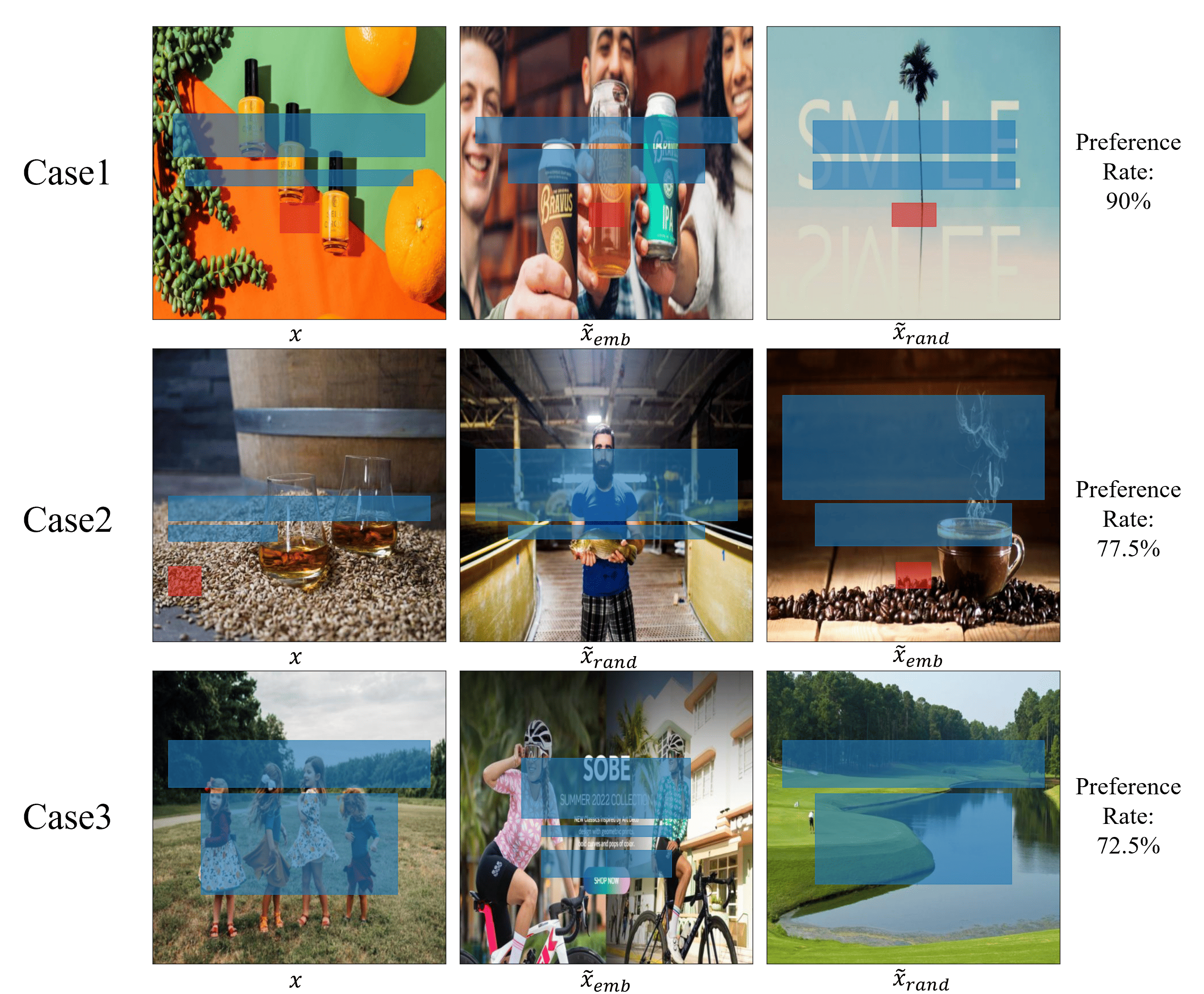}
    \caption{User study examples.
    Each row shows one triplet: reference $x$ (left), embedding-based retrieval $\tilde{x}_{\text{emb}}$ (middle),
    and random retrieval $\tilde{x}_{\text{rand}}$ (right).
    The \textit{Preference Rate} on the right shows how often $\tilde{x}_{\text{emb}}$ was chosen by participants.
    Blue bounding boxes denote text regions, while red bounding boxes denote button elements.}

    \label{supp:fig:user_study}
\end{figure*}

Figure~\ref{supp:fig:user_study} shows three representative cases,
with the \textit{Preference Rate} on the right indicating how often
$\tilde{x}_{\text{emb}}$ was chosen over $\tilde{x}_{\text{rand}}$.
In Case~1, $\tilde{x}_{\text{emb}}$ was preferred 90\% of the time,
in Case~2 the rate was 77.5\%, and in Case~3 it was 72.5\%.
Across all 30 triplets, $\tilde{x}_{\text{emb}}$ was chosen in 83.4\% of comparisons,
demonstrating that our learned embeddings align well with human perception.
These results also suggest that the proposed TemplateFID metric is consistent
with human judgments of template similarity.

\subsection{Dataset}

We conduct our experiments on the Web-design dataset~\cite{weng2024desigen}, which consists of around 50K web banner designs collected from real-world online shopping platforms. Each sample includes a high-resolution background image, structured layout annotations (bounding boxes and element types), and a product description that conveys information such as commercial purpose, theme, or target audience. We use 41,270 samples (85\%) for training, 2,427 samples (5\%) for validation, and 4,856 samples (10\%) for testing.

\subsection{Implementation Details of the InterIL Model}
\label{supp:sec:dualflow_details}

\noindent \textbf{Layout Prior.} Each layout is represented as a sequence of up to 7 elements, where each element is defined by five attributes: category, left coordinate, top coordinate, width, and height. We discretize each attribute into 64 values by replacing raw coordinates with their nearest k-means cluster index. Layouts with fewer than 7 elements are padded to a fixed length. Notably, only about 0.97\% of layouts in the dataset contain more than 7 elements.

\noindent \textbf{Image Prior.} For a fair comparison between our method, OpenCOLE, and Desigen, we use the same SD-v1.4 backbone and set the output resolution to $512 \times 512$, consistent with Desigen. For the teaser in the supplementary material, we adopt SD-v2.0 with an output resolution of $768 \times 768$.

We use a $\beta$-VAE to model the layout, with a latent size of 32. The VAE encodes layout tokens into compact latent representations. For layout denoising, we adopt a transformer-based architecture using DiT as the backbone. Specifically, our Layout LDM consists of 28 transformer blocks, each with 8 attention heads and a query/key/value dimension of 512. This configuration enables the model to effectively capture spatial dependencies and hierarchical structures across layout elements during the denoising process.

\noindent \textbf{Communication Module.} In the joint modeling stage, we introduce a communication module that facilitates feature exchange between the layout and image backbones. Specifically, we extract multi-scale image features from the second and third downsampling blocks as well as the middle block of the image U-Net, and use the output features of the 19th transformer block of the layout diffusion model as the layout representation. Cross-attention is performed in both directions: the layout representation attends to the image features to incorporate visual context, while the image features attend to the layout representation to integrate structural information. The image-informed layout representation is fused with the original 19th-block output through a residual connection and then fed into the 20th transformer block; analogously, the attended image representations are fused back into the corresponding U-Net blocks. During this stage, both the pre-trained layout diffusion model and image backbone are kept frozen, and only the communication module is updated.

\noindent \textbf{Inference Details.} During inference, we use a DDIM sampler for the layout backbone and a DDPM sampler for the image backbone, each with 50 sampling steps. The communication module is enabled at timesteps greater than 700 (corresponding to $\rho=30\%$ of the denoising trajectory) to facilitate cross-modal information exchange. In addition, both occlusion-aware guidance (OAG) and readability-aware guidance (RAG) are applied between steps 200 and 700.

\subsection{Training Details}
\label{supp:sec:training_details}
Our training pipeline consists of two main stages: (1) \textbf{Building Domain-specific Priors}, which includes pretraining the layout VAE, training the Layout LDM, and fine-tuning Stable Diffusion; and (2) \textbf{Training the Joint Model}, which focuses on learning the cross-domain communication module. All experiments are conducted on an NVIDIA A40 GPU with 48 GB of memory.

\paragraph{Stage 1: Building Domain-specific Priors.}
\begin{itemize}[leftmargin=*]
    \item \textit{Layout VAE Training.} We employ a $\beta$-VAE~\cite{higgins2017beta} to encode layouts into latent representations. The $\beta$ coefficient is set to $5 \times 10^{-4}$ to find a sweet spot between reconstruction accuracy and latent disentanglement.

    \item \textit{Layout LDM Training.} The layout diffusion model is trained to model the distribution of latent layouts. We use a batch size of 4096 and train for 1000 epochs with a learning rate of $1 \times 10^{-4}$.

    \item \textit{Stable Diffusion Fine-tuning.} To ensure fair comparison with \textbf{Desigen}~\cite{weng2024desigen}, the previous state-of-the-art method, we fine-tune the Stable Diffusion model on the corresponding dataset. The model is trained for 100 epochs with a batch size of 8 and a learning rate of $1 \times 10^{-5}$.
\end{itemize}

\paragraph{Stage 2: Training the Joint Model.}
\begin{itemize}[leftmargin=*]
    \item \textit{Cross-Domain Communication Module.} The feature exchange module is trained independently to facilitate information transfer between the layout and image domains. The model is trained for 20 epochs with a batch size of 20 and a learning rate of $1 \times 10^{-4}$.
\end{itemize}

\paragraph{Optimization.}
All training stages adopt the AdamW optimizer with $\beta_1 = 0.9$, $\beta_2 = 0.999$, and a weight decay of $1 \times 10^{-2}$. The learning rate schedule follows a cosine decay strategy.

\begin{figure*}[htbp]
\centering
\includegraphics[width=0.9\linewidth]{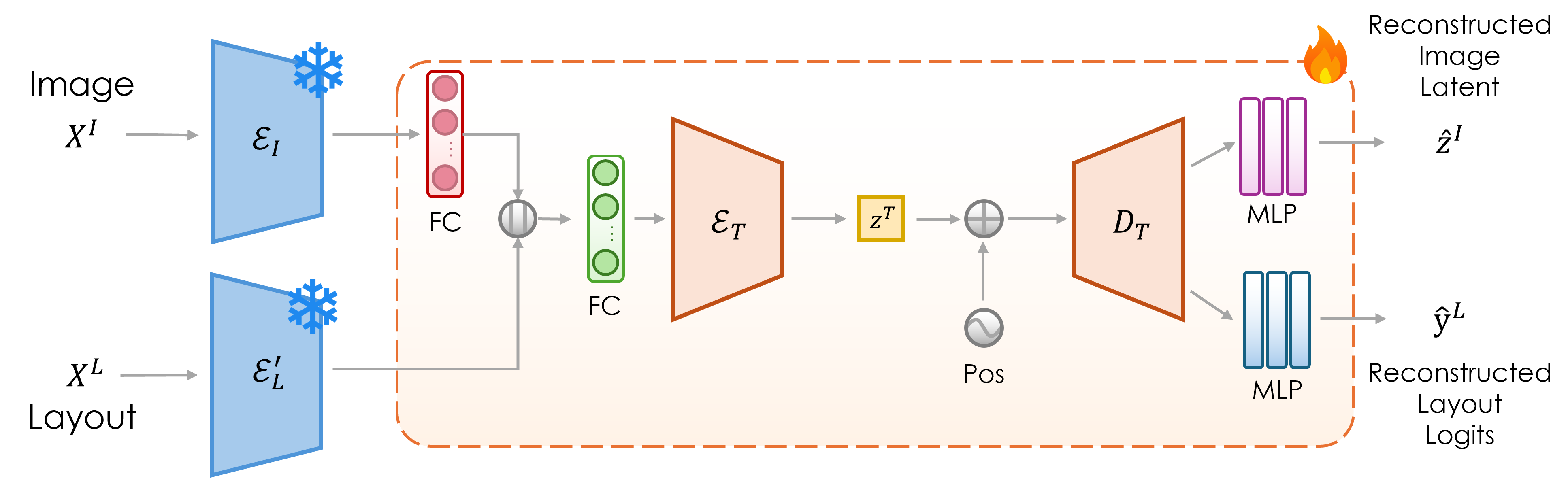}
\caption{Architecture of the template autoencoder (TemplateAE). The frozen image encoder $\mathcal{E}_{\text{I}}$ and the independent TemplateAE layout encoder $\mathcal{E}_{\text{L}}^{'}$ map the image and layout to their latent spaces. The independent layout encoder $\mathcal{E}_{\text{L}}^{'}$ is trained on the mixture dataset and is not shared with InterIL. The concatenated embeddings are processed by the template encoder $\mathcal{E}^T$ to produce a unified design latent $z^T$, which is then decoded by the template decoder $\mathcal{D}^T$ and two MLPs to reconstruct the image latent and layout logits.}

\label{supp:fig:dt-ae}
\end{figure*}

\subsection{Implementation Details of Preference-Specific Guidance}
We now detail the two guidance objectives used in the paper. Both are plugged into the same inference-time guidance framework and differ only in the choice of differentiable objective computed from the decoded image-layout predictions.

\noindent \textbf{Occlusion-Aware Guidance (OAG).} Given the predicted image noise $\hat{\epsilon}_t^I$, we compute a clean image latent $\hat{z}_0^I$ analytically using the forward-process marginal distribution $q(z_t^I|z_0^I)$, decode it into an image $\mathcal{D}_{\text{I}}(\hat{z}_0^I)$, and compute a saliency map $S = f_\text{sal}(\mathcal{D}_{\text{I}}(\hat{z}_0^I))$ using an off-the-shelf saliency detector $f_\text{sal}$. We then compute a clean layout latent $\hat{z}_0^L$ from the predicted layout noise $\hat{\epsilon}_t^L$ with $q(z_t^L|z_0^L)$, and decode it into a layout $\mathcal{D}_{\text{L}}(\hat{z}_0^L)$. Let $\{\mathbf{b}_i\}_{i=1}^B$ be the decoded element bounding boxes. The OAG objective is written as:
\begin{align}
    \mathcal{L}_\text{occ} = \frac{1}{B}\sum_{i=1}^B \frac{1}{A_i} \sum_{p} M_{\mathbf{b}_i}(p)\, S(p),
\end{align}
where $M_{\mathbf{b}_i}$ is a soft mask for $\mathbf{b}_i$, $A_i$ is the area of $\mathbf{b}_i$, and $p$ indexes spatial positions.

Since the decoded layout elements have discrete bounding box parameters, we convert them into continuous ones to make the guidance objectives differentiable. We estimate each continuous parameter as a weighted sum of quantization-bin centers, where the weights are the predicted probabilities over bins. We denote by $\mathbf{b}_i = (x_i, y_i, w_i, h_i) \in \mathbb{R}^4$ the estimated continuous bounding box parameters of the $i$-th element, where $(x_i, y_i)$ denotes the top-left location, and $w_i$ and $h_i$ are, respectively, the width and height. From $\mathbf{b}_i$, we compute the top-left coordinates $(x^l_i, y^t_i)$ and the bottom-right coordinates $(x^r_i, y^b_i)$ of the bounding box. The soft mask $M_{\mathbf{b}_i}$ is computed in a differentiable manner, with the value at position $(x, y)$:
\begin{equation}
\begin{split}
    M_{\mathbf{b}_i}(x, y) =  \sigma & (\lambda (x - x^l_i)) \times \sigma(\lambda (x^r_i - x)) \\
        & \times \sigma(\lambda (y - y^t_i)) \times \sigma(\lambda (y^b_i - y)),
\end{split}
\end{equation}
where $\sigma(\cdot)$ is the sigmoid function. $\lambda$ is set to 40. This differentiable box representation is used in both OAG and RAG.

\noindent \textbf{Readability-Aware Guidance (RAG).} For RAG, we use the same decoded image $\mathcal{D}_{\text{I}}(\hat{z}_0^I)$ and decoded layout $\mathcal{D}_{\text{L}}(\hat{z}_0^L)$, but replace the saliency map with a differentiable clutter map $C$ computed from spatial gradient operators on the decoded image, so that regions with stronger edges, textures, or other high-frequency patterns receive larger values. From the decoded layout, we select the subset of boxes corresponding to text elements, denoted by $\{\mathbf{b}^{\text{text}}_i\}_{i=1}^{B_T}$. The RAG objective is written as:
\begin{align}
    \mathcal{L}_{\text{read}} = \frac{1}{B_T}\sum_{i=1}^{B_T} \frac{1}{A_i} \sum_{p} M_{\mathbf{b}^{\text{text}}_i}(p)\, C(p),
\end{align}
where $M_{\mathbf{b}^{\text{text}}_i}$ is the soft mask for the $i$-th decoded text box. In this way, OAG and RAG share the same inference-time mechanism and differ only in their preference-specific objective functions.

\subsection{TemplateAE Architecture Details}

To enable holistic evaluation of design templates, we train a template autoencoder (TemplateAE) to extract joint image-layout embeddings. For TemplateFID, TemplateAE is trained on a large-scale mixture dataset comprising GenPoster100K~\cite{wang2025sega}, CGL~\cite{ijcai2022p692}, Crello~\cite{yamaguchi2021canvasvae}, and PKU~\cite{Hsu-2023-posterlayout}, deliberately excluding Web-design to avoid evaluation bias.

Before training TemplateAE, we separately train another Layout VAE on the same mixture dataset and use its encoder, denoted by $\mathcal{E}_{\text{L}}^{'}$, as the frozen layout encoder of TemplateAE. This Layout VAE is independent of the one used by InterIL and shares neither parameters nor training data with it.

As shown in Figure~\ref{supp:fig:dt-ae}, given a background image $X^I$ and layout $X^L$, they are first mapped to latent space through the frozen image encoder $\mathcal{E}_{\text{I}}$ and the independent layout encoder $\mathcal{E}_{\text{L}}^{'}$ to obtain $z^I = \mathcal{E}_{\text{I}}(X^I)$ and $z^L = \mathcal{E}_{\text{L}}^{'}(X^L)$. These embeddings are concatenated and processed through a template encoder $\mathcal{E}^T$ to produce a unified design latent $z^T$. Subsequently, $z^T$ is combined with cosine position embeddings and fed into a template decoder $\mathcal{D}^T$, followed by two separate MLPs that reconstruct the respective domain outputs: image latent $\hat{z}^I$ and layout logits $\hat{y}^L \in \mathbb{R}^{N \times d^L \times V}$. During training, we freeze $\mathcal{E}_{\text{I}}$ and $\mathcal{E}_{\text{L}}^{'}$ and only optimize $\mathcal{E}^T$, $\mathcal{D}^T$, and the MLPs. The training objective on this mixture dataset combines reconstruction losses for both modalities:

\begin{equation}
    \mathcal{L}_{\text{rec}} = \gamma\big(1-\frac{z^I \cdot \hat{z}^I}{\|z^I\| \|\hat{z}^I\|}\big) + \sum_{i=1}^{N} \sum_{j=1}^{V} -X^L_{i,j} \log(\hat{y}^L_{i,j})
\end{equation}

where the first term measures image latent reconstruction quality and the second term evaluates layout prediction accuracy. $\gamma$ is the balance factor that brings the two terms to comparable scales; in our experiments, we set $\gamma=5.0$. For TemplateCLIP, we initialize from this mixture-dataset-pretrained TemplateAE encoder and further fine-tune it on the Web-design train split together with the pretrained SD text encoder using a CLIP-style contrastive objective.

\subsection{Further Inference Time Analysis}
\label{supp:sec:inference_time}

All inference times reported in Table~\ref{supp:tab:merged_inference_time} are measured on a
\textbf{single NVIDIA A40 GPU} for fair and consistent comparison.
Varying the communication ratio $\rho$ has only a minor effect on inference time, which decreases from 5.24\,s at $\rho=100\%$ to 4.91\,s at $\rho=0\%$.

Table~\ref{supp:tab:merged_inference_time} summarizes the average time required to generate one
design template across different models on the Web-design test set.
Our joint diffusion framework achieves substantially faster generation
than both \textbf{Desigen}~\cite{weng2024desigen} and \textbf{OpenCOLE}~\cite{jia2023cole},
while maintaining superior design quality.

Specifically, our base model requires only \textbf{5.01\,s} per template on average,
achieving a \textbf{1.5$\times$--6$\times$ speed-up} over Desigen variants,
whose multi-stage iterative refinement introduces considerable latency at each iteration.
Even with guidance variants such as \textbf{OAG} and \textbf{RAG}, or with GPT-augmented prompts,
our variants remain competitive (8.69\,s, 8.47\,s, and 10.14\,s, respectively),
and still significantly outperform OpenCOLE (25.8\,s).
This efficiency primarily stems from our \textit{single-pass joint denoising process},
which eliminates redundant layout--image alternation and avoids the sequential generation bottlenecks
present in prior frameworks.

In contrast, \textbf{OpenCOLE} adopts a more complex three-stage pipeline:
a \textit{design plan generation module} expands the input text;
an \textit{image generation module} synthesizes a background image;
and a \textit{typography module} predicts layout elements.
While this decomposition improves controllability, it substantially increases inference latency
because these components must be executed sequentially.
Consequently, OpenCOLE requires \textbf{25.8\,s} per template, approximately
\textbf{5$\times$ slower} than our unified framework.

Overall, these results demonstrate that \textbf{InterIL} not only captures image--layout interactions
effectively, but also scales efficiently for real-time or large-scale design generation scenarios.

\begin{table}[htbp]
\caption{
Average inference time per design template on the Web-design test set.
Our joint framework is substantially faster than Desigen and OpenCOLE, while the guided variants add only modest overhead.
}
\label{supp:tab:merged_inference_time}
\centering
\footnotesize
\setlength{\tabcolsep}{4pt}
\renewcommand{\arraystretch}{1.15}
\begin{tabular}{@{}lc@{}}
\toprule
\textbf{Method} & \textbf{Time (s / template)} \\
\midrule
Desigen (0) & 7.47 \\
Desigen (1) & 14.85 \\
Desigen (2) & 22.42 \\
Desigen (3) & 29.89 \\
OpenCOLE & 25.80 \\
\midrule
\textbf{Ours} & \textbf{5.01} \\
Ours + OAG & 8.69 \\
Ours + RAG & 8.47 \\
Ours (Prompt Aug.) & 10.14 \\
\bottomrule
\end{tabular}
\end{table}

\end{document}